\documentclass[iicol]{sn-jnl}

\jyear{2021}%

\theoremstyle{thmstyleone}%
\theoremstyle{thmstyletwo}%

\theoremstyle{thmstylethree}%

\usepackage{amsmath,amsfonts}
\usepackage{array}
\usepackage{subfigure}
\usepackage{textcomp}
\usepackage{stfloats}
\usepackage{url}
\usepackage{verbatim}
\usepackage{graphicx}
\usepackage{cite}
\usepackage{multirow}
\usepackage{booktabs}
\usepackage{setspace}
\usepackage{xcolor}
\usepackage{url}
\usepackage{bbm}
\usepackage{pdfpages}
\newcommand{\hc}[1]{{\color{black} #1}}
\newcommand{\rhc}[1]{{\color{black} #1}}
\newcommand{\mr}[1]{{\color{black} #1}}
\begin{document}

\title[Article Title]{Hierarchical Skeleton Meta-Prototype Contrastive Learning with Hard Skeleton Mining for Unsupervised Person Re-Identification}


\author[1,2]{\fnm{Haocong} \sur{Rao}}\email{haocong001@ntu.edu.sg}
\author[2,3]{\fnm{Cyril} \sur{Leung}}\email{cleung@ece.ubc.ca}
\author*[1,2]{\fnm{Chunyan} \sur{Miao}}\email{ascymiao@ntu.edu.sg}


\affil[1]{\orgdiv{School of Computer Science and Engineering}, \orgname{Nanyang Technological University}, \orgaddress{\country{Singapore}}}

\affil[2]{\orgdiv{Joint NTU-UBC Research Centre of Excellence in Active Living for the Elderly (LILY)}, \orgname{Nanyang Technological University}, \orgaddress{\country{Singapore}}}


\affil[3]{\orgdiv{Department of Electrical and Computer Engineering}, \orgname{The University of British Columbia}, \orgaddress{\country{Canada}}}



\abstract{With rapid advancements \hc{in} depth sensors and deep learning, skeleton-based person re-identification (re-ID) \mr{models} have recently \hc{achieved} remarkable progress with many advantages. Most existing solutions learn single-level skeleton features from body joints with the assumption of equal skeleton importance, while they typically lack the ability to exploit more informative skeleton features \hc{from various levels such as limb level with more global body patterns}. The label dependency of these methods also limits their flexibility in learning more general skeleton representations. This paper proposes a generic unsupervised Hierarchical skeleton Meta-Prototype Contrastive learning (Hi-MPC) approach with Hard Skeleton Mining (HSM) for person re-ID with \textit{unlabeled} 3D skeletons. 
Firstly, we construct hierarchical representations of skeletons to \hc{model \textit{coarse-to-fine} body and motion features from the levels of body joints, components, and limbs.} Then a hierarchical meta-prototype contrastive learning model is proposed to cluster and contrast the most typical skeleton features (\textit{“prototypes”}) from different-level skeletons. By converting original prototypes into \textit{meta-prototypes} with multiple homogeneous transformations, we \hc{induce} the model to learn the inherent consistency of prototypes to capture more effective skeleton features for person re-ID. Furthermore, we devise a hard skeleton mining mechanism to adaptively infer the informative importance of each skeleton, so as to focus on harder skeletons to learn more discriminative skeleton representations. Extensive evaluations on five datasets demonstrate that our approach outperforms a wide variety of state-of-the-art skeleton-based methods. 
We further show the general applicability of our method to cross-view person re-ID and RGB-based scenarios with estimated skeletons.}




\maketitle

\section{Introduction}
Person re-identification (re-ID) is \hc{a pattern recognition task that aims to retrieve and match} a certain pedestrian across different views or occasions. Recent years have witnessed its great success in many applications such as intelligent video surveillance, security authentication, human tracking, and robotics
\cite{nambiar2019gait,zheng2015towards,baltieri2011sarc3d,vezzani2013people,su2018multi,chen2018person,li2018semi,li2019unsupervised,qian2019leader,lan2020semi,yu2020unsupervised,wu2020rgb,ye2021deep}.
With \hc{the} surging popularity of low-cost, non-invasive, and accurate skeleton-tracking sensors ($e.g.,$ Kinect \cite{shotton2011real-time}), person re-ID via 3D skeletons has attracted increasing attention in both academia and industry
\cite{munaro20143d,munaro2014one,barbosa2012re,andersson2015person,pala2019enhanced,liao2020model,rao2020self,rao2021self,rao2021multi,rao2021sm,rao2022simmc,rao2023transg,rao2022revisiting}, \mr{while many important topics of skeleton-based person re-ID ($e.g.,$ how to improve both accuracy and scalability) still remain to be studied} \cite{rao2022simmc,rao2022skeleton}. In contrast to conventional methods that utilize visual human appearances and body textures from RGB or depth images \cite{wang2003silhouette,wang2011human,wang2016person,zhao2017person,zhang2019densely,karianakis2018reinforced,ge2020selfpaced}, skeleton-based person re-ID models \hc{provide numerous advantages} such as \mr{(1) smaller data size with concise 3D human body representation ($i.e.$, 3D coordinates of key body joints), (2) better privacy protection \textit{without} using appearance information, (3) more robust performance under scale, view, and background variations \cite{han2017space,rao2021self}}.

Traditional skeleton-based methods \cite{barbosa2012re,munaro20143d,munaro2014one,andersson2015person,pala2019enhanced} typically extract hand-crafted features like skeleton descriptors in terms of pre-defined anthropometric and gait attributes of body. Nevertheless, these methods \rhc{rely heavily} on domain expertise \rhc{such as} anatomy and kinematics  \cite{yoo2002extracting} to model skeleton data, \rhc{and} lack the flexibility to fully exploit latent features beyond human cognition. Recent mainstream methods \cite{liao2020model,rao2020self,rao2021multi,rao2021self,rao2021sm,rao2022simmc,rao2023transg} resort to deep neural networks (DNNs) to perform \hc{skeleton representation learning.}
Despite the great efforts, they \hc{usually} require manually-annotated skeleton data to train or fine-tune models, which is labor-expensive and could reduce their general applicability in practice. 
Another crucial \hc{shortcoming} of these methods is that they typically learn skeleton features from a single level ($e.g.,$ body joint level \cite{rao2020self,rao2021self}) and assume that each skeleton is equally important in representing \hc{the} patterns of a person \cite{rao2021self,rao2022simmc}. \rhc{This} intrinsically limits their ability to exploit key features of more informative skeletons. For instance, there usually exist skeletons that are either harder to be recognized as the same identity ($i.e.,$ large intra-class variations) or easier to be misidentified among different individuals ($i.e.,$ small inter-class variations), both of which should be \hc{given} greater attention for learning more effective skeleton representations.  

\begin{figure}
    \centering
    \scalebox{0.6}{
    \includegraphics{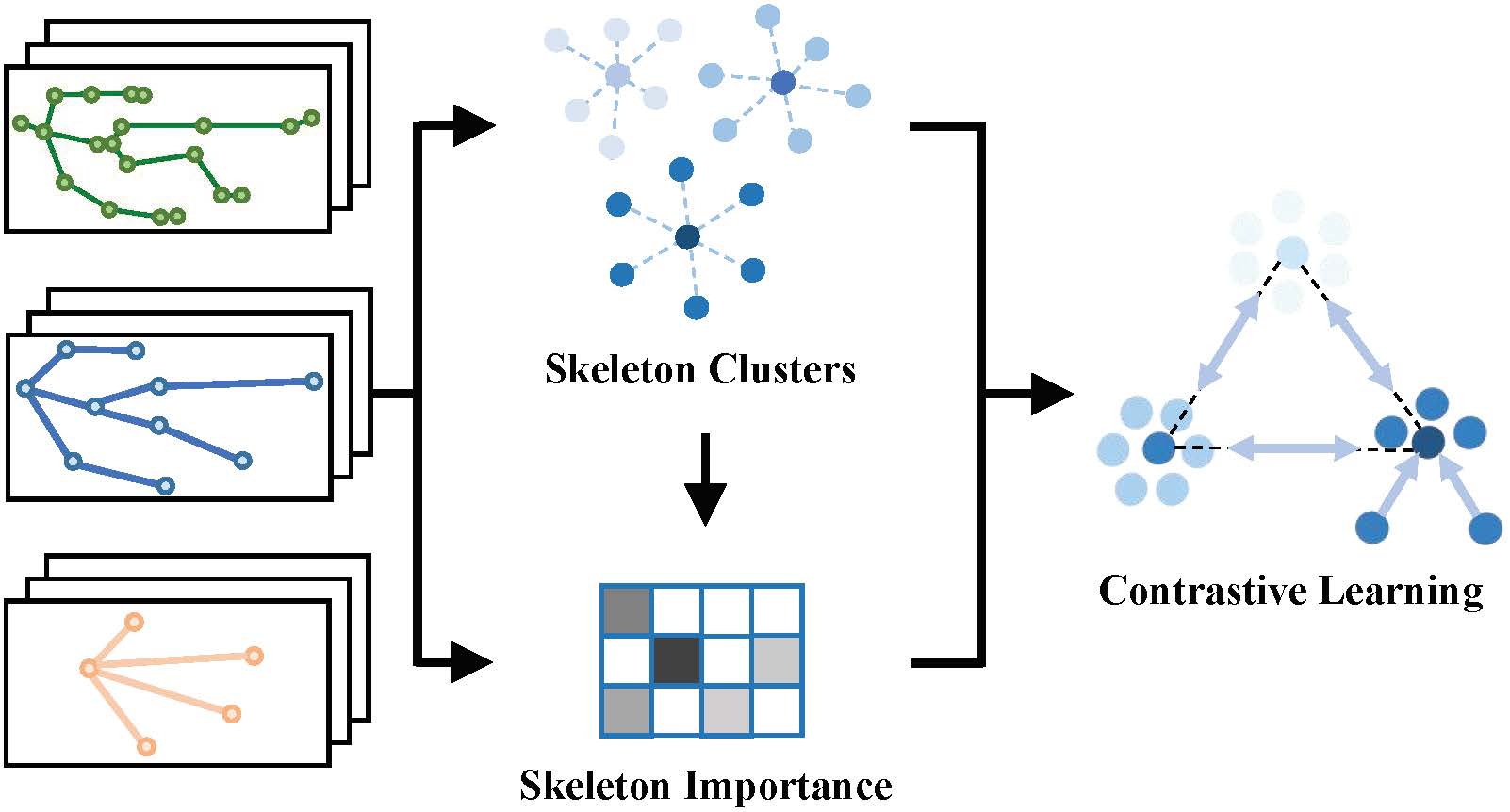}
    }
    \caption{Our approach hierarchically clusters skeleton representations to infer their inherent importance, and contrasts the key clustered features with the most typical ones to \mr{learn effective skeleton representations for unsupervised person re-ID.}}
    \label{first}
\end{figure}

To solve the aforementioned challenges for \textit{unsupervised skeleton-based} person re-ID, we propose a generic Hierarchical skeleton Meta-Prototype Contrastive learning (Hi-MPC) approach with a Hard Skeleton Mining (HSM) mechanism, as shown in Fig. \ref{first}. \hc{The approach} exploits \textit{unlabeled} hierarchical representations of key informative skeletons to contrast and learn the most typical skeleton features for person re-ID.
Firstly, we construct a hierarchical representation for each skeleton with coarse-to-fine body partitions, which enables the model to explore body structures and pattern information from different levels. Secondly, a Hierarchical skeleton Meta-Prototype Contrastive learning (Hi-MPC) approach is devised to cluster and contrast the most representative skeleton features (\hc{defined} as \textit{“prototypes”}) from \textit{different-level} skeleton representations (\hc{defined} as \textit{“instances”}). To encourage learning more consistent and representative skeleton prototypes, we propose to transform original instances into multiple \textit{homogeneous meta-instances}, and maximize their inherent similarity to corresponding \textit{meta-prototypes} while maximizing their dissimilarity to others, so as to capture discriminative skeleton features and class-related semantics ($e.g.,$ intra-class similarity) from various levels of unlabeled skeletons. Thirdly, \hc{considering that different skeletons usually possess varying informative value, $e.g.,$ some skeletons are more difficult to be classified to the correct identity (defined as \textit{“hard skeletons”}) but can provide more informative clues for model learning \cite{hermans2017defense}, we \textit{for the first time} devise a Hard Skeleton Mining (HSM) mechanism to adaptively infer the importance of each skeleton in learning hard and easily-confused patterns. In this way, HSM enables our model to mine and focus on hard skeletons in Hi-MPC to encourage more effective skeleton representation learning.} Lastly, we propose to construct the novel Multi-level Skeleton Meta-Representation (MSMR) that combines skeleton features learned from different levels as the final representation for person re-ID. As a byproduct of our approach, we reveal the feasibility of exploiting more concise and abstract skeleton representations to perform person re-ID.

Our contributions are summarized as follows:
\begin{itemize}
\item We devise hierarchical representations of 3D skeletons and propose a novel hierarchical skeleton meta-prototype contrastive learning approach with a hard skeleton mining mechanism to learn effective representations from \textit{unlabeled} skeleton sequences for person re-ID.

\item We propose the Hierarchical skeleton Meta-Prototype Contrastive learning (Hi-MPC) that hierarchically contrasts representative features and inherent similarity of different-level skeleton representations to learn discriminative features and high-level semantics for person re-ID.

\item We devise a Hard Skeleton Mining (HSM) mechanism to adaptively infer informative importance of skeletons within each sequence to encourage learning more effective skeleton representations from harder skeletons.

\item We empirically validate the effectiveness of each level skeleton representation learned from the proposed approach, and combine them to construct the novel Multi-level Skeleton Meta-Representation (MSMR) for person re-ID.

\item \hc{Extensive experiments on five public benchmarks demonstrate that our approach outperforms most state-of-the-art methods on person re-ID tasks.} We further show that our method is generally effective in multi-view and RGB-based scenarios with estimated skeletons.

\end{itemize}

The rest of \hc{this} paper is organized as follows. Sec. \ref{related} introduces relevant works on skeleton-based person re-identification and contrastive learning. Sec. \ref{approach} \hc{describes} technical components of the proposed approach. Sec. \ref{experiments} provides experimental details and a comprehensive comparison between our approach and existing \hc{state-of-the-art methods}. Sec. \ref{discussion} details \hc{our} ablation study and empirical analysis of the proposed approach. Sec. \ref{conclusion} concludes this paper. 

\section{Related Works}
\label{related}
\subsection{Skeleton-Based Person Re-Identification}
\subsubsection{Hand-Crafted Methods} 
Most conventional methods manually extract 3D skeleton features from anthropometric and gait aspects to depict human body and motion patterns \cite{barbosa2012re,andersson2015person,munaro2014one,pala2019enhanced}. In \cite{barbosa2012re}, seven Euclidean distances between certain joints are integrated into a learnable distance matrix for person re-ID.
They are further extended into 13 (\mr{denoted as $D_{13}$}) and 16 skeleton descriptors (\mr{denoted as $D_{16}$}) in \cite{munaro2014one} and \cite{pala2019enhanced} respectively, which are learned by different classifiers ($k$-nearest neighbor, support vector machine or \hc{Adaboost}) to perform person re-ID tasks. Due to the limited performance of existing descriptors and their inherent requirement of domain expertise, these methods are usually combined with more efficient features such as 3D point clouds \cite{munaro20143d} or face descriptors \cite{pala2019enhanced} to boost accuracy. \mr{In our work, three representative hand-crafted descriptors ($i.e.,$ non-model-based methods), including $D_{13}$ \cite{munaro2014one}, $D_{16}$ \cite{pala2019enhanced}, and the PoseGait descriptor (denoted as $D_{\text{PG}}$) in \cite{liao2020model} that combines pose features, joint angles, limb lengths, and joint motion are compared with deep learning based models.}

\subsubsection{Deep Learning Based Methods} 
Recent years \hc{have} witnessed the great success of deep learning in supervised and self-supervised skeleton representation learning \cite{liao2020model,rao2020self,rao2021self,rao2021multi,rao2021sm,rao2022simmc}. In \cite{liao2020model}, a CNN-based architecture PoseGait is leveraged to learn 81 pre-defined skeleton/pose features (\mr{$D_{\text{PG}}$}) for supervised human recognition. A \hc{self-supervised} skeleton encoding model (AGE) with locality-aware attention based LSTM \cite{rao2020self} is devised to encode discriminative gait patterns for person re-ID. Its extension 
SGELA \cite{rao2021self} further enhances self-supervised skeleton semantic learning with diverse skeletal pretext tasks ($e.g.,$ time series forecasting \cite{feng2022multi}) and inter-sequence contrastive mechanisms for the person re-ID task. \rhc{Skeleton} graphs are constructed in \cite{rao2021multi} based on the physical connections of body joints or parts to learn skeletal relations and high-level motion semantics via graph attention networks \cite{feng2022relation} for person re-ID. In \cite{rao2021sm}, multi-scale skeleton reconstruction and cross-scale skeleton inference are further integrated into a graph encoding framework for self-supervised person re-ID.
The SimMC framework \cite{rao2022simmc} performs single-level prototype contrastive learning and intra-sequence similarity learning with randomly masked skeleton sequences to realize unsupervised person re-ID. \mr{The skeleton/pose data are also utilized in many image/video-based person re-ID methods to help extract pre-defined local/hierarchical body parts \cite{su2017pose,wei2017glad}, disentangle semantic components ($e.g.,$ bone locations) \cite{wang2022pose,lu2023exploring} or generate pose augmented representations \cite{liu2018pose}. The key differences between these methods and our approach are two-fold. First, they use skeleton data as \textit{auxiliary} information to combine with RGB images to boost the model performance, while our work exploits only unlabeled skeleton data without using any appearance-based features for person re-ID. Second, they typically leverage the hierarchical structure of the original body skeletons to generate image-based body parts, our approach hierarchically constructs skeleton representations at different levels, each of which corresponds to a new \textit{independent} body representation to learn discriminative pattern information for person re-ID.
}

Compared to previous hand-crafted, supervised, and self-supervised \mr{skeleton-based} models, \hc{our} approach requires no pre-defined skeleton descriptors, graph modeling or pretext design for skeleton representation learning. Our unsupervised method can efficiently exploit \textit{unlabeled} 3D skeleton data to hierarchically mine the most \hc{representative} features from different-level skeleton representations, which enables capturing comprehensive body patterns and coarse-to-fine skeleton semantics for person re-ID tasks. \hc{This work is the first exploration} of potential hard skeletons to exploit more informative patterns, and it can be seamlessly integrated into unsupervised contrastive learning to enhance discriminative feature mining. 

\subsection{Contrastive Learning}
Contrastive learning has recently driven various self-supervised and unsupervised paradigms \cite{wu2018unsupervised,oord2018representation,he2020momentum,chen2020a,rao2021self,chen2021exploring,li2021prototypical,xiao2022learning} to learn efficient data representations in a way of pulling together positive representation pairs and pushing apart negative ones in a certain feature space. An instance discrimination method with exemplar tasks and noise-contrastive estimation (NCE) \cite{gutmann2010noise} is proposed in \cite{wu2018unsupervised} for visual contrastive learning. A context auto-encoding task with \hc{probabilistic} contrastive loss, InfoNCE, is utilized in the contrastive predictive coding (CPC) model \cite{oord2018representation} to learn effective representations from different domains. The mini-batch negative sampling mechanism \cite{chen2020a} and momentum-based encoder \cite{he2020momentum} are devised to improve the capacity and consistency of contrastive paradigms, while a Siamese architecture is further explored in \cite{chen2021exploring} to perform contrastive representation learning without using negative pairs or momentum encoders.
In \cite{li2021prototypical}, contrastive learning and \textit{k}-means clustering are integrated into a framework for unsupervised visual learning. 

Fundamentally different from previous studies that leverage augmented samples of images as contrastive instances, \hc{our} work devises a new generic meta-prototype contrastive learning paradigm for 3D skeleton data, and exploits unlabeled hierarchical skeleton representations as instances to mine the most representative and discriminative features ($i.e.,$ skeleton prototypes) for instance-prototype contrastive learning. In contrast to existing skeleton contrastive learning methods \cite{rao2021self,rao2022simmc}, we explore the hierarchical contrastive learning of multi-level skeleton representations, and propose the meta-transformation of instances and prototypes to enhance the inherent consistency and effectiveness of prototype learning. Our approach requires neither extra sampling scheme nor memory mechanism, while it can fully exploit skeletal importance to enhance the contrastive learning of harder and more informative samples to achieve more effective skeleton representations.

\mr{
\subsubsection{Hard Negative Mining for Contrastive Learning}
Hard negative mining aims to find more informative training samples that are difficult to discriminate ($e.g.,$ easily-confused negatives), which has been widely applied to various areas to accelerate network training and improve model performance  \cite{hermans2017defense,ge2018deep,hu2021adco,wang2021instance,kalantidis2020hard,verma2021towards,zhang2022m,robinson2021contrastive,jeon2021mining}. In contrastive learning, existing hard negative mining models can be mainly grouped into (1) \textit{adversarial learning based methods} \cite{hu2021adco,wang2021instance} that generate hard negatives by adversarial optimization and (2) \textit{mixing based methods} \cite{kalantidis2020hard,verma2021towards,zhang2022m} that mix the negative samples and positive samples in the feature space. In \cite{hu2021adco}, a representation network and its negative adversary are alternately trained to generate the hardest negative samples for contrasting, while \cite{wang2021instance} further introduces a diversity loss to generate diverse challenging negative samples with different noise. 
A hard negative mixing strategy is proposed in \cite{kalantidis2020hard} to mix features of hardest negatives and its query to enhance contrastive learning. To improve the difficulty of generated negatives, a diversity objective function is devised in \cite{zhang2022m} to mix multiple samples with dynamic weights. Some recent works also explore controllable hard negative mining with an importance sampling strategy \cite{robinson2021contrastive} or dynamic curriculum learning \cite{jeon2021mining}. 

Different from previous methods that require extra triplet constraint \cite{schroff2015facenet,hermans2017defense}, adversarial learning \cite{hu2021adco,wang2021instance} or sample mixing \cite{kalantidis2020hard,verma2021towards,zhang2022m}, the proposed hard skeleton mining mechanism can directly exploit the inherent similarity between skeleton representations (meta-instance) and cluster-level representations (meta-prototypes) to adaptively infer the informative importance of skeletons at different skeleton levels and different feature subspaces, and mine 
 both hard positive and negative skeleton samples from sequences without using any labels. Our approach is specifically designed to mine key informative skeletons within each sequence, which can be generally applied to different level skeleton representations to enhance the proposed hierarchical contrastive learning for unsupervised skeleton-based person re-ID tasks.
}

\begin{figure*}
    \centering
    \scalebox{0.595}{\includegraphics{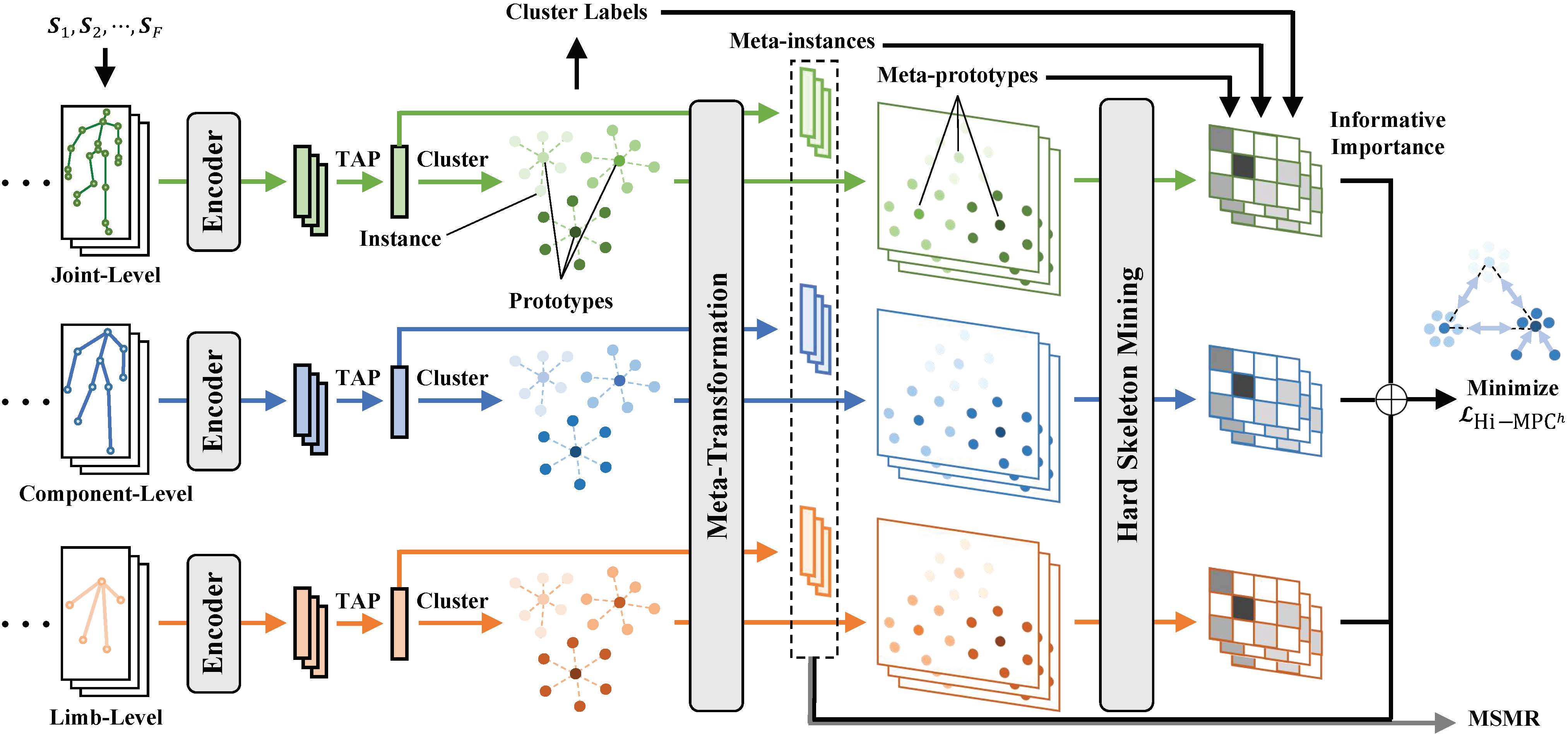}}
    \caption{Schematics of our approach. Firstly, each skeleton sequence $\boldsymbol{S}_{1},\boldsymbol{S}_{2},\cdots,\boldsymbol{S}_{F}$ is hierarchically represented at \hc{joint-level (top), component-level (middle), and limb-level (bottom)}. Secondly, we perform temporal average pooling (TAP) on
    the encoded skeleton representations of each level to generate skeleton instances, and cluster them to find prototypes, which are then transformed into meta-instances and meta-prototypes in different contrastive subspaces. Lastly, a hard skeleton mining mechanism (illustrated in Fig. \ref{HSM}) is employed to infer informative importance of skeletons within each sequence, which is integrated into the contrastive loss $\mathcal{L}_{\text {Hi-MPC}^{h}}$ to enhance the similarity of key meta-instances belonging to the same meta-prototype while maximizing their dissimilarity to other meta-prototypes. The meta-instances learned from different levels are combined to construct multi-level skeleton meta-representation (MSMR) for the person re-ID task.}
    \label{overview}
\end{figure*}

\section{The Proposed Approach}
\label{approach}
The goal of our approach is to perform \textit{unsupervised} person re-identification with unlabeled 3D skeleton sequences. Formally, we denote a sequence of 3D skeletons as $\boldsymbol{S}_{1:F}\!=\!(\boldsymbol{S}_1,\cdots,\boldsymbol{S}_{F})\in \mathbb{R}^{F \times K}$, where $\boldsymbol{S}_{i}\in \mathbb{R}^{K}$ is the $i^{th}$ skeleton with 3D positions of $J$ body joints and $K=3\times J$.
The training set $\Phi^{\mathcal{T}}=\left\{\boldsymbol{S}^{ \mathcal{T},i}_{1:F}\right\}_{i=1}^{N_{1}}$, probe set $\Phi^{\mathcal{P}}=\left\{\boldsymbol{S}^{ \mathcal{P},i}_{1:F}\right\}_{i=1}^{N_{2}}$, and gallery set $\Phi^{\mathcal{G}}=\left\{\boldsymbol{S}^{\mathcal{G}, i}_{1:F}\right\}_{i=1}^{N_{3}}$ contain $N_{1}$, $N_{2}$, and $N_{3}$ skeleton sequences of different persons in varying views or occasions. Each skeleton sequence $\boldsymbol{S}_{1:F}$ corresponds to a unique identity $\text{y}\in \{1, \cdots, I\}$ where $I$ is the number of different identities. 
Our approach aims at learning to encode $\Phi^{\mathcal{P}}$ and $\Phi^{\mathcal{G}}$ into effective skeleton representations $\{\boldsymbol{V}^{\mathcal{P}}_i\}_{i=1}^{N_{2}}$ and $\{\boldsymbol{\boldsymbol{V}}^{\mathcal{G}}_j\}_{j=1}^{N_{3}}$ \textit{without using any label}, such that the representation $\boldsymbol{V}^{\mathcal{P}}_i$ in the probe set can match the representation $\boldsymbol{V}^{\mathcal{G}}_j$ of the same identity in the gallery set. 
The overview of our approach is presented in Fig. \ref{overview}, and we detail each technical component below.

\subsection{Hierarchical Skeleton Representations}
\label{hier_rep_sec}
\hc{The} human body can be naturally modeled with several key functional regions at different levels ($e.g.,$ joints, limbs) \cite{winter2009biomechanics,rao2021multi}, which could \textit{hierarchically} characterize different anthropometric or kinetic features of \hc{the} body. Inspired by this fact, \hc{we spatially divide each human skeleton into various non-overlapping partitions, each of which corresponds to a certain body part at the level of joints, body components ($e.g.$, hands) or limbs ($e.g.$, upper limbs). Then we generate the position of each body part by computing the centroid of body joints within the corresponding partition.} As presented in Fig. \ref{overview}, we build hierarchical skeleton representations for each skeleton $\boldsymbol{S}$ from three levels, namely \textit{joint-level}, \textit{component-level}, and \textit{limb-level} skeleton representations, which can correspondingly contain low, middle, and high level body structures and pattern information of a skeleton. Formally, the $l^{th}$ level representation $\boldsymbol{\hat{S}}^{l}\in \mathbb{R}^{3 \times n_{l}}$ ($l\in\{1,2,3\}$) consists of 3D positions of $n_{l}$ body partitions, where $n_{1}=J$, $n_{2}=10$, $n_{3}=5$ correspond to joint-level ($\boldsymbol{\hat{S}}^{1}$), component-level ($\boldsymbol{\hat{S}}^{2}$), limb-level skeleton representations ($\boldsymbol{\hat{S}}^{3}$), respectively. 
We denote the hierarchical representations of each input skeleton sequence $\boldsymbol{S}_{1:F}$ as $\boldsymbol{\hat{S}}^{1}_{1:F}\in \mathbb{R}^{F\times 3n_{1}}$, $\boldsymbol{\hat{S}}^{2}_{1:F}\in \mathbb{R}^{F\times 3n_{2}}$, and $\boldsymbol{\hat{S}}^{3}_{1:F} \in \mathbb{R}^{F\times 3n_{3}}$.

\subsection{Hierarchical Skeleton Meta-Prototype Contrastive Learning}
\label{Hi-HPC_sec}
As each pedestrian's skeletons typically possess unique identity-associated features in terms of anthropometric attributes ($e.g.,$ body part lengths) and walking patterns \cite{murray1964walking}, it is desirable to exploit the \textit{most typical skeleton features (``prototypes'')} from skeleton sequences \textit{(``instances'')} to differentiate \hc{a given} person from others. \mr{A straightforward solution is to find the skeleton prototypes, which can represent a unique skeleton attribute or concept belonging to certain identities, by clustering skeleton instances for direct prototype-instance contrastive learning in a \textit{single} feature space \cite{rao2022simmc}.}
\hc{However}, the inherent randomness of feature initialization or clustering \cite{li2021prototypical} could induce unstable prototype estimation and inconsistent relational distributions ($e.g.,$ prototype-instance relations) when representation spaces vary.
\rhc{Based on the assumption} that \hc{the} global distribution of prototypes should be consistent with the distribution of cluster centroids (\hc{referred to as} \textit{“prototype-cluster consistency”}, see Appendix \uppercase\expandafter{\romannumeral1}), we propose to construct different contrastive subspaces that \rhc{inherit} from the original feature space of prototypes to enhance contrastive learning. In particular, we perform prototype-instance contrasting in each \textit{individual} contrastive subspace, which are combined based on the prototype-cluster consistency to encourage more robust probability estimation of prototypes and more consistent contrastive learning. 
\hc{To this end,} we devise the \textit{\textbf{hierarchical skeleton meta-prototype contrastive learning (Hi-MPC)}} to homogeneously transform original prototypes and instances into \textit{\textbf{meta-prototypes}} and \textit{\textbf{meta-instances}} at each skeleton level, and contrast their inherent similarity in different transformed contrastive subspaces to \textit{jointly} learn representative discriminative skeleton features for person re-ID. 


Given the $l^{th}$ level skeleton representations $\boldsymbol{\hat{S}}^{l}_{1},\cdots,\boldsymbol{\hat{S}}^{l}_{F}$ of an input skeleton sequence, we first encode them and apply temporal average pooling (TAP) to obtain a sequence-level skeleton representation, $i.e.,$ instance (shown in Fig. \ref{overview}) as:
\begin{equation}
\boldsymbol{v}^{l,(i)}=\frac{1}{F}\sum^{F}_{j=1}\psi^{l}\left(\boldsymbol{\hat{S}}^{l,(i)}_{j}\right)=\frac{1}{F}\sum^{F}_{j=1}\boldsymbol{z}^{l,(i)}_{j},
\label{eq_1}
\end{equation}
where $\psi^{l}(\cdot)$ is the $l^{th}$ level encoder built by a multi-layer perceptron (MLP) network with one hidden layer, $\boldsymbol{z}^{l,(i)}_{j}\in\mathbb{R}^{h_1}$ denotes the encoded features of the $l^{th}$ level representation of $j^{th}$ skeleton in the $i^{th}$ training skeleton sequence, and $\boldsymbol{v}^{l,(i)}\in\mathbb{R}^{h_1}$ denotes the encoded $l^{th}$ level representation of $i^{th}$ training skeleton sequence $\boldsymbol{\hat{S}}^{l,(i)}_{1:F}, i\in\{1, \cdots,N_1\}$. Here we adopt TAP to \textit{average} the temporal dynamics of all skeletons to represent the features of a sequence \cite{rao2022simmc}. It is worth noting that TAP also keeps the consistency of feature dimensions between skeleton-level and sequence-level representations. \rhc{This} allows us to directly compute their inherent similarity by dot products without extra dimension transformation (see Sec. \ref{hsm_sec}).


Then, to mine \hc{the} original skeleton prototypes, we exploit the encoded sequence-level representations $\mathbb{V}^{l}=\{\boldsymbol{v}^{l,(1)},\cdots,\boldsymbol{v}^{l,(N_1)}\}$ as skeleton instances, and leverage the DBSCAN algorithm \cite{ester1996density} to cluster instances of similar features and semantics. As shown in Fig. \ref{overview}, we generate clusters as $\widehat{\mathbb{V}}_{c}^{l}=\{\boldsymbol{v}^{l}_{c,k}\}_{k=1}^{n_{c}}$, $c\in\{1,\cdots,C\}$, where $C$ denotes the number of different clusters, $i.e.,$ pseudo classes, and the $c^{th}$ cluster $\widehat{\mathbb{V}}_{c}^{l}$ contains $n_c$ instances. Note that we perform clustering \textit{individually} on each level of hierarchical skeleton representations to better capture different level semantics and retain coarse-to-fine skeleton features.
The instance features of the same cluster are \hc{averaged} as the corresponding skeleton prototype with:                                                   
\begin{equation}
\boldsymbol{p}^{l}_{c}=\frac{1}{n_c}\sum^{n_c}_{k=1}\boldsymbol{v}^{l}_{c,k} \ ,
\label{eq_2}
\end{equation}
where $\boldsymbol{p}^{l}_{c}\in\mathbb{R}^{h_1}$ denotes the original skeleton prototype of the $c^{th}$ cluster $\widehat{\mathbb{V}}_{c}^{l}$ generated from the $l^{th}$ level skeleton instances.
Given \hc{the} original prototypes and instances, our model converts them into \textit{meta-prototypes} and \textit{meta-instances} with multiple meta-transformation heads by: 
\begin{equation}
(\boldsymbol{\hat{v}}^{l}_{c,k})^{m}= \mathbf{H}^{l,m}_{1} \boldsymbol{v}^{l}_{c,k}\ ,
\label{eq_3}
\end{equation}
\begin{equation}
(\boldsymbol{\hat{p}}^{l}_{c})^{m}=\mathbf{H}^{l,m}_{2} \boldsymbol{p}^{l}_{c}\ ,
\label{eq_4}
\end{equation}
where $(\boldsymbol{\hat{v}}^{l}_{c,k})^{m}, (\boldsymbol{\hat{p}}^{l}_{c})^{m} \in \mathbb{R}^{h_2}$ denote the $m^{th}$ meta-instance and meta-prototype transformed from $\boldsymbol{v}^{l}_{c,k}$ and $\boldsymbol{p}^{l}_{c}$. Here $\mathbf{H}^{l,m}_{1},\mathbf{H}^{l,m}_{2}\in\mathbb{R}^{h_2\times h_1}$ are corresponding learnable weight matrices of the $m^{th}$ transformation head. $m\in\{1,\cdots,M\}$ and $M$ denotes the number of different transformation heads. 
Considering that both original instances and prototypes come from the same domain, $i.e.,$ being represented in the \textit{homogeneous} feature space of the same dimension \cite{DBLP:journals/ai/ZhouPT19}, it is natural to employ homogeneous feature mapping for each pair of heads (\hc{defined as} \textit{“meta-transformation heads”}) with $\mathbf{H}^{l,m}_{1}=\mathbf{H}^{l,m}_{2}$ and $h=h_1=h_2$. The meta-transformation heads map both instances and prototypes into the \textit{same} $m^{th}$ new feature space, which can be viewed as the $m^{th}$ subspace \textit{linearly} transformed from the original contrastive feature space, to generate homogeneous meta-instances and meta-prototypes. It should be noted that we do NOT use \textit{heterogeneous} feature mapping ($i.e.,$ $\mathbf{H}^{l,m}_{1}\neq\mathbf{H}^{l,m}_{2}$), as it separately maps instances and prototypes into two different feature subspaces with domain shifts \cite{sun2016return} and \rhc{degrades} the model performance (see Appendix \uppercase\expandafter{\romannumeral2}).

To jointly focus on representative skeleton features \hc{of} all meta-prototypes and \hc{capture} different-level skeleton semantics ($e.g.,$ class-related patterns) from \hc{diverse} contrastive feature subspaces, we propose the Hi-MPC loss below:
\begin{equation}
\scalebox{0.81}{
$
\begin{aligned}
    \mathcal{L}_{\text {Hi-MPC}}=\sum_{l=1}^{3}\sum_{i=1}^{I_l}\sum_{m=1}^{M}-\log \frac{\exp \left((\boldsymbol{\hat{v}}^{l, (i)})^{m} \cdot (\boldsymbol{\hat{p}}^{l}_{+})^{m} / \tau \right)}{\sum_{c=1}^{C} \exp \left((\boldsymbol{\hat{v}}^{l,(i)})^{m} \cdot (\boldsymbol{\hat{p}}^{l}_{c})^{m} / \tau \right)},
    \label{eq_5}
\end{aligned}
$
}
\end{equation}
where $I_l$ denotes the number of instances in all clusters generated from the $l^{th}$ level skeleton representations, $(\boldsymbol{\hat{v}}^{l,(i)})^{m}$ denotes the $m^{th}$ transformed meta-instance of $i^{th}$ instance belonging to its corresponding meta-prototype $(\boldsymbol{\hat{p}}^{l}_{+})^{m}$,  $(\boldsymbol{\hat{p}}^{l}_{c})^{m}$ is the meta-prototype of the $c^{th}$ cluster at the $l^{th}$ level, and $\tau$ is the temperature for contrastive learning. We set $\tau=\sqrt{h}$ to scale the dot products to improve the stability of contrastive learning \cite{vaswani2017attention}. Note that $\mathcal{L}_{\text {Hi-MPC}}$ is averaged over all meta-instances for training.
The Hi-MPC approach combining both hierarchical skeleton clustering (see Eq. (\ref{eq_1}) and (\ref{eq_2})) and \hc{multiple} meta-transformations (see Eq. (\ref{eq_3}) and (\ref{eq_4})) enables our model to perform a coarse-to-fine skeleton prototype estimation and mine different-level skeleton semantics ($e.g.,$ identity-specific semantics), and also encourages more consistent prototype learning by jointly attending to key meta-prototypes in different representation subspaces. However, Hi-MPC only considers sequence-level skeleton representations, $i.e.,$ instances \hc{with} averaged skeleton features (see Eq. (\ref{eq_1})), and cannot 
fully exploit key skeletons with higher importance in each sequence for contrastive learning, which motivates us to propose the hard skeleton mining mechanism below.

\begin{figure}
    \centering
    \scalebox{0.82}{
    \includegraphics{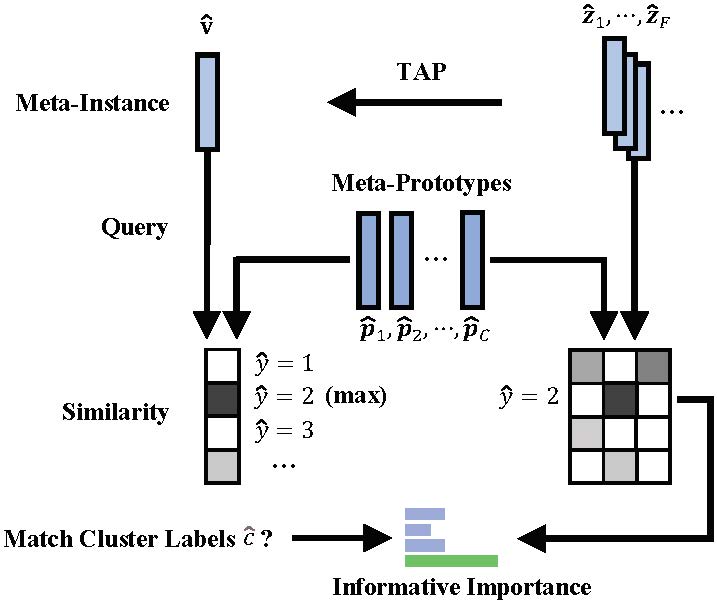}
    }
    \caption{Computation flow of HSM mechanism. The similarity between skeleton representations $\boldsymbol{\hat{z}}_{1},\cdots,\boldsymbol{\hat{z}}_{F}$ and the meta-prototype ($\hat{\text{y}}$) predicted by the  meta-instance $\boldsymbol{\hat{v}}$ is first queried. The informative importance is \hc{then} inferred based on the true or false matching of the cluster label $\hat{c}$.}
    \label{HSM}
\end{figure}

\subsection{Hard Skeleton Mining Mechanism}
\label{hsm_sec}
Different skeletons within the same sequence could \hc{possess different} importance (\textit{\hc{referred to as} ``\textbf{informative importance}''}) \hc{in mining hard ($e.g.,$ easily-confused) patterns of a person.} In particular, similar-looking skeletons and patterns shared among different \hc{persons} (\textit{\hc{referred to as} ``hard negatives''}), or wildly different poses of the same person (\textit{\hc{referred to as} ``hard positives''}), are typically harder to be distinguished while they can provide more informative clues for models to comprehend the full concept of ``same person'' \cite{hermans2017defense}. To achieve this goal, we propose the \textit{\textbf{hard skeleton mining ($\boldsymbol{HSM}$) mechanism}} to encourage the model to focus on skeleton representations with higher informative importance from hard negatives and hard positives in Hi-MPC.

As shown in Fig. \ref{HSM}, \hc{given the skeleton sequence representation $\boldsymbol{\hat{v}}$ that belongs to the $\hat{c}^{th}$ cluster, we first predict its label} by querying the dot product based similarity with meta-prototypes in the $m^{th}$ contrastive feature subspace by:
\begin{equation}
   \hat{\text{y}}^{m}=\underset{i}{\arg \max } \left(\boldsymbol{(\hat{v}})^{m} \cdot (\boldsymbol{\hat{p}}_{i})^{m}\right).
    \label{eq_6}
\end{equation}
\hc{In Eq. (\ref{eq_6}),} $\hat{\text{y}}^{m} \in \{1, \cdots, C\}$ is the predicted cluster label, $(\boldsymbol{\hat{v}})^{m}$ and $(\boldsymbol{\hat{p}}_{i})^{m}$ denote the meta-instance and the $i^{th}$ meta-prototype generated by the $m^{th}$ meta-transformation head. It is worth noting that the \textit{cluster label} (denoted as $\hat{c}$) generated by DBSCAN algorithm \hc{is adopted as} the ground-truth label since the real label is not available. We use the label of \rhc{the} cluster centroid ($i.e.,$ meta-prototype) that has the maximum similarity with $(\boldsymbol{\hat{v}})^{m}$ as the predicted cluster label of $(\boldsymbol{\hat{v}})^{m}$ in the $m^{th}$ contrastive feature subspace. For convenience, in Eq. (\ref{eq_6}) we omit the superscripts of levels and use $\boldsymbol{\hat{S}}_{1:F}$, $(\boldsymbol{\hat{v}})^{m}$, and $(\boldsymbol{\hat{p}}_{i})^{m}$ to denote the $l^{th}$ level representation $\boldsymbol{\hat{S}}^{l}_{1:F}$, $(\boldsymbol{\hat{v}}^{l})^{m}$, and $(\boldsymbol{\hat{p}}_{i}^{l})^{m}$, while $\hat{\text{y}}^{m}$ corresponds to the predicted cluster label of the meta-instance $(\boldsymbol{\hat{v}})^{m}$.

\begin{algorithm*}[h] 
\setstretch{1.3}
	\caption{Main Algorithm of Hi-MPC with HSM} 
	\label{algorithm_1}
	\begin{algorithmic}[1] 
		\Require 
		Unlabeled training skeleton sequences $\Phi^{\mathcal{T}}=\left\{\boldsymbol{S}^{\mathcal{T}, (i)}_{1:F}\right\}_{i=1}^{N_{1}}$, initialized encoder function $\psi(\cdot)$, initialized $M$ meta-transformation heads $\operatorname{Meta}^{m}(\cdot)$, temperature $\tau$
		\Ensure 
		Encoder $\psi(\cdot)$, meta-transformation head $\operatorname{Meta}^{m}(\cdot)$
		\State $\boldsymbol{\hat{S}}^{1,(i)}_{1:F}, \boldsymbol{\hat{S}}^{2,(i)}_{1:F}, \boldsymbol{\hat{S}}^{3,(i)}_{1:F} = \operatorname{Hier}\left(\boldsymbol{S}^{\mathcal{T},(i)}_{1:F}\right)$
		\begin{footnotesize}\Comment{Hierarchical skeleton representations at joint/component/limb-level}\end{footnotesize} 
		\Repeat

    \State $\boldsymbol{v}^{l,(i)}=\operatorname{TAP}\left( \psi(\boldsymbol{\hat{S}}^{l,(i)}_{1:F})\right)= \operatorname{TAP}\left({\boldsymbol{z}^{l,(i)}_{1},\cdots,\boldsymbol{z}^{l,(i)}_{F}}\right)$\begin{footnotesize}\Comment{Encode hierarchical skeleton sequences into instances}\end{footnotesize}

    \State $\{\widehat{\mathbb{V}}_{c}^{l}\}_{c=1}^{C}=\operatorname{DBSCAN}\left(\{\boldsymbol{v}^{l,(i)}\}_{i=1}^{N_{1}}\right)$\begin{footnotesize}\Comment{Find clusters and discard outliers} \end{footnotesize}

     \State $\boldsymbol{p}_{c}^{l}=\operatorname{Proto}\left(\widehat{\mathbb{V}}_{c}^{l}\right)$ \begin{footnotesize}\Comment{Generate skeleton prototypes with Eq.(\ref{eq_2})}\end{footnotesize}
     

    \State $\left((\boldsymbol{\hat{v}}^{l}_{c,k})^{m},(\boldsymbol{\hat{p}}^{l}_{c})^{m}\right)=\operatorname{Meta}^{m}\left(\boldsymbol{v}^{l}_{c,k},\boldsymbol{p}^{l}_{c}\right)$\begin{footnotesize}\Comment{Perform the $m^{th}$ meta-transformation with Eq.(\ref{eq_3}), (\ref{eq_4})} \end{footnotesize}

    \State $\hat{\text{y}}^{m}=\operatorname{Pred}\left(({\hat{z}}^{l})^{m}\right)=\operatorname{Pred}\left(({\hat{v}}^{l})^{m}\right)$\begin{footnotesize}\Comment{Predict cluster label for $({\hat{v}}^{l})^{m}$ and its skeletons with Eq.(\ref{eq_6})} \end{footnotesize} 
    
    \State Use predicted meta-prototype $(\boldsymbol{\hat{p}}_{\hat{\text{y}}^{m}})^{m}$ to infer importance $\overline{\delta}\left(({\hat{z}}^{l})^{m}\right)$ of each skeleton with Eq. (\ref{eq_8})

    \State $ \mathcal{L}_{\text     {Hi-MPC}^{h}}\left(\overline{\delta}\left((\boldsymbol{\hat{z}}_{j}^{l})^{m}\right),(\boldsymbol{\hat{z}}^{l}_{j})^{m},\{(\boldsymbol{\hat{p}}^{l}_{c})^{m}\}^{C}_{c=1}, \tau\right)$\begin{footnotesize}\Comment{Compute importance-weighted contrastive loss with Eq.(\ref{eq_9})} \end{footnotesize}  
    
    \State Update parameters of $\psi(\cdot)$ and $\operatorname{Meta}^{m}(\cdot)$ to minimize $\mathcal{L}_{\text{Hi-MPC}^{h}}$
    \Until{\textit{MaxEpoch} or \textit{MaxPatience}} 
	\end{algorithmic} 
\end{algorithm*}

\hc{Having obtained} encoded features $\boldsymbol{z}_{1},\cdots,\boldsymbol{z}_{F}$ of $F$ skeletons in $\boldsymbol{\hat{S}}_{1:F}$, where $\boldsymbol{z}_{j}=\psi\left(\boldsymbol{\hat{S}}_{j}\right)$ and $j \in \{1,\cdots,F\}$  (see Eq. (\ref{eq_1})), we assign the cluster label $\hat{\text{y}}^{m}$ predicted by their sequence-level representation $(\boldsymbol{\hat{v}})^{m}$ to each of them.
As illustrated in Fig. \ref{HSM}, we compute the inherent similarity of each skeleton representation to the predicted cluster \hc{using}:
\begin{equation}
   \delta((\boldsymbol{\hat{z}}_{j})^{m}) = \frac{\exp \left((\boldsymbol{\hat{z}}_{j})^{m} \cdot (\boldsymbol{\hat{p}}_{\hat{\text{y}}^{m}})^{m}\right)}{\sum_{t=1}^{F} \exp \left((\boldsymbol{\hat{z}}_{t})^{m} \cdot (\boldsymbol{\hat{p}}_{\hat{\text{y}}^{m}})^{m} \right)}.
    \label{eq_7}
\end{equation}
\hc{In Eq. (\ref{eq_7}),} $\delta((\boldsymbol{\hat{z}}_{j})^{m}) \in (0, 1)$ represents the \textit{normalized} similarity between the representation of $j^{th}$ skeleton and the meta-prototype corresponding to the predicted $\hat{\text{y}}^{m}$ cluster in the $m^{th}$ contrastive subspace. For clarity and consistency, we use $(\boldsymbol{\hat{z}}_{j})^{m}$ to represent the $j^{th}$ skeleton representation transformed by the $m^{th}$ head corresponding to Eq. (\ref{eq_3}).
$\delta((\boldsymbol{\hat{z}}_{j})^{m})$ can be interpreted as the the degree of certainty that the $j^{th}$ skeleton within the sequence is classified to $\hat{\text{y}}^{m}$ in the $m^{th}$ feature subspace, while higher certainty indicates that the skeleton is easier for learning to realize correct classification. Hence, the informative importance of each skeleton in the same sequence can be inferred by:
\begin{equation}
   \overline{\delta}((\boldsymbol{\hat{z}}_{j})^{m}) =  \frac{\mathbb \exp \left(\mathbb I(\hat{\text{y}}^{m}, \hat{c})\left((\boldsymbol{\hat{z}}_{j})^{m} \cdot (\boldsymbol{\hat{p}}_{\hat{\text{y}}^{m}})^{m}\right)\right)}{\sum_{t=1}^{F} \exp \left(\mathbb I(\hat{\text{y}}^{m}, \hat{c})\left((\boldsymbol{\hat{z}}_{t})^{m} \cdot (\boldsymbol{\hat{p}}_{\hat{\text{y}}^{m}})^{m} \right)\right)},
    \label{eq_8}
\end{equation}
where $\overline{\delta}((\boldsymbol{\hat{z}}_{j})^{m})\in (0, 1)$ represents the informative importance of the $j^{th}$ skeleton in the sequence $\boldsymbol{\hat{S}}_{1:F}$ when being represented in the $m^{th}$ contrastive feature subspace, and
$\mathbb I(\hat{\text{y}}^{m}, \hat{c})=-1$ if the predicted label $\hat{\text{y}}^{m}$ of $(\boldsymbol{\hat{z}}_{j})^{m}$ is $\hat{c}$ otherwise $\mathbb I(\hat{\text{y}}^{m}, \hat{c})=1$. Intuitively, when the label prediction is true, $i.e.,$ $\hat{\text{y}}^{m}=\hat{c}$, the hardest positive is the skeleton with the \textit{lowest} certainty ${\delta}(\cdot)$, which is more likely to contain diverse patterns of the same person and possesses higher informative importance, thus we have $\mathbb I(\hat{\text{y}}^{m}, \hat{c})=-1$ and $\overline{\delta}(\cdot)\propto \frac{1}{{\delta}(\cdot)}$. When the model fails to predict correctly, $i.e.,$ $\hat{\text{y}}^{m}\neq \hat{c}$, the hardest negative is the most certain skeleton being classified to the false label, while it contains more similar information that needs to be carefully distinguished. In this case, we have $\overline{\delta}(\cdot)\propto {\delta}(\cdot)$, which is naturally achieved with $\mathbb I(\hat{\text{y}}^{m}, \hat{c})=1$. To facilitate coarse-to-fine pattern learning, the proposed HSM is performed on each level of skeleton hierarchical representations.  

To fully exploit skeletons within each sequence and focus on harder skeletons with higher informative importance for Hi-MPC training, we integrate the skeleton importance into contrastive learning by proposing the Hi-MPC$^{h}$ loss as follows:
\begin{equation}
\scalebox{0.81}{
$
\begin{aligned}
    &\mathcal{L}_{\text {Hi-MPC}^{h}}=\\
    &\sum_{l=1}^{3}\sum_{i=1}^{I_l}\sum_{j=1}^{F}\sum_{m=1}^{M}\overline{\delta}\left((\boldsymbol{\hat{z}}_{j}^{l,(i)})^{m}\right) \operatorname{Softmax}_{j}\left((\boldsymbol{\hat{z}}^{l, (i)}_{j})^{m} \cdot (\boldsymbol{\hat{p}}^{l}_{+})^{m} / \tau\right),
    \label{eq_9}
\end{aligned}
$
}
\end{equation}
where $\operatorname{Softmax}_{j}\left((\boldsymbol{\hat{z}}^{l, (i)}_{j})^{m} \cdot (\boldsymbol{\hat{p}}^{l}_{+})^{m} / \tau\right)= -\log \frac{\exp \left((\boldsymbol{\hat{z}}^{l, (i)}_{j})^{m} \cdot (\boldsymbol{\hat{p}}^{l}_{+})^{m} / \tau \right)}{\sum_{c=1}^{C} \exp \left((\boldsymbol{\hat{z}}^{l, (i)}_{j})^{m} \cdot (\boldsymbol{\hat{p}}^{l}_{c})^{m} / \tau \right)}$. The proposed $\mathcal{L}_{\text {Hi-MPC}^{h}}$ in Eq. (\ref{eq_9}) inherits from $\mathcal{L}_{\text {Hi-MPC}}$ (see Eq. (\ref{eq_5})) and combines the proposed HSM mechanism to adaptively infer the informative importance $\overline{\delta}(\cdot)$ of $F$ skeletons in each sequence to enhance the proposed hierarchical skeleton meta-prototype contrastive learning. Instead of directly leveraging sequence-level skeleton representations $\boldsymbol{v}^{l,(i)}$ for Hi-MPC (see Sec. \ref{Hi-HPC_sec}), the proposed $\mathcal{L}_{\text {Hi-MPC}^{h}}$ can take advantage of finer-grained  pattern information contained in \textit{each key skeleton} and corresponding hierarchical representations to \mr{enhance the Hi-MPC learning (Eq. (\ref{eq_5}))} to mine more discriminative skeleton features for person re-ID tasks. \hc{More details about theoretical and empirical analyses are provided in the appendices.}

\begin{figure}[t]
    \centering
         \scalebox{0.215}{\includegraphics{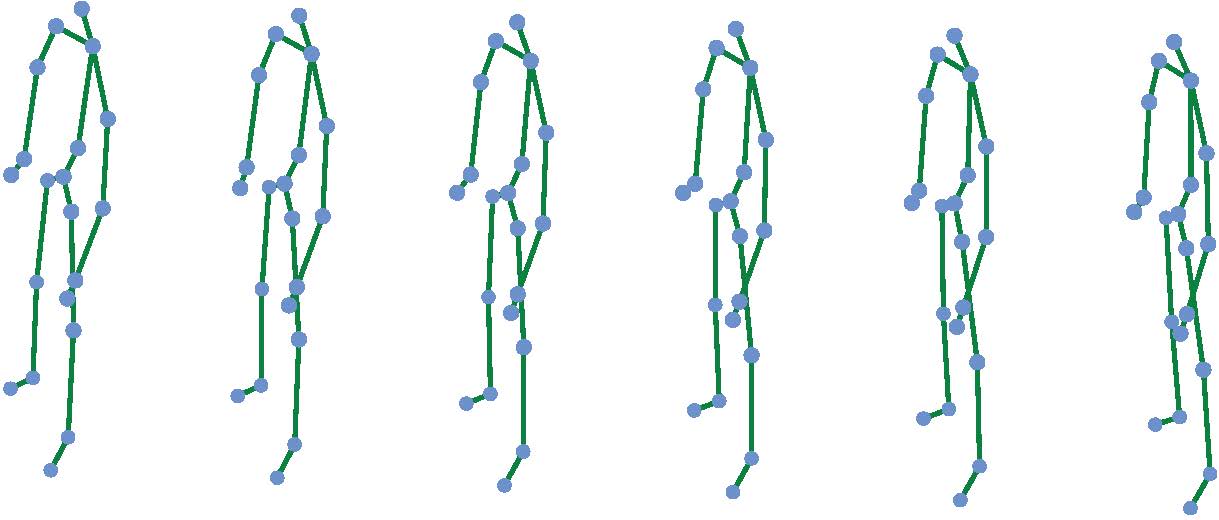}} 
           \quad         
            \scalebox{0.245}{\includegraphics{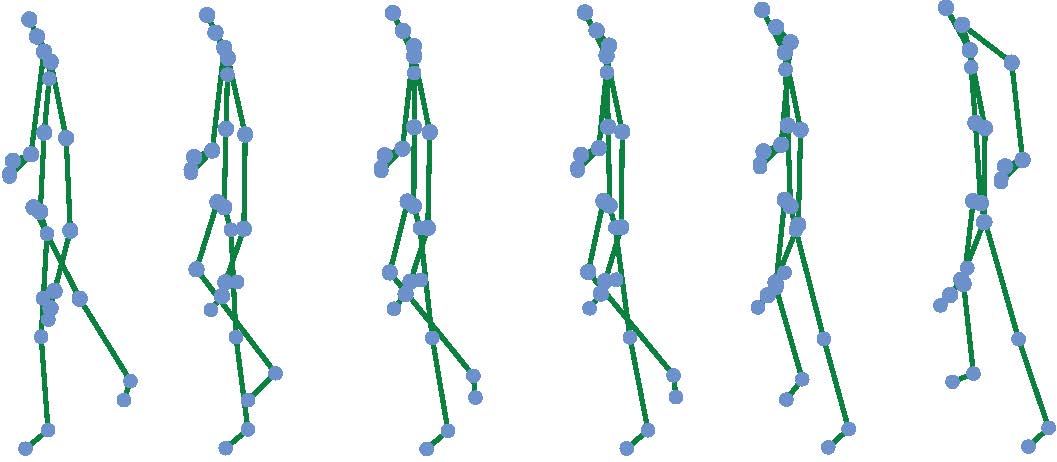}} 
           \scalebox{0.255}{\includegraphics{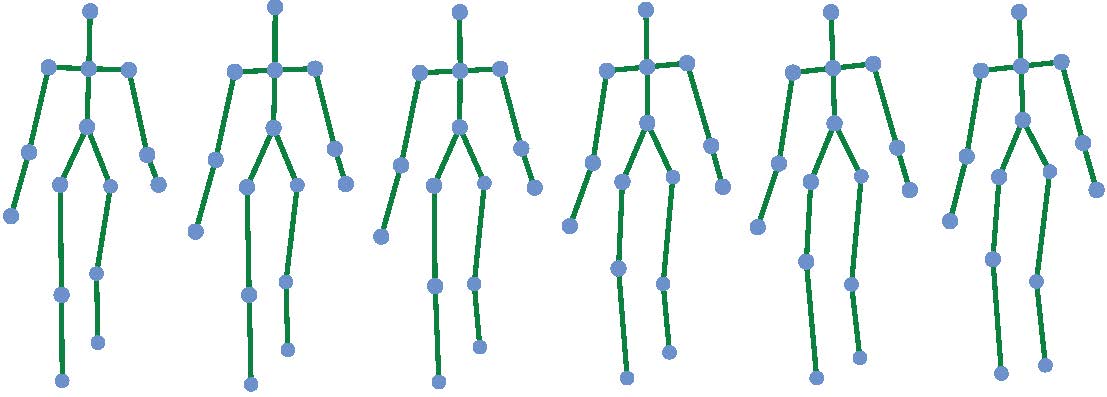}}   
          \
       \scalebox{0.11}{\includegraphics{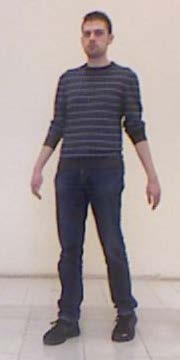}}
       \scalebox{0.11}{\includegraphics{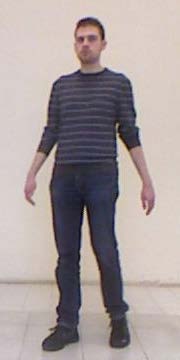}}
        \scalebox{0.11}{\includegraphics{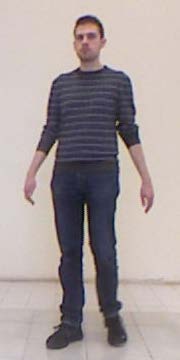}}
         \scalebox{0.11}{\includegraphics{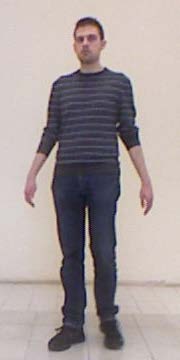}}
          \scalebox{0.11}{\includegraphics{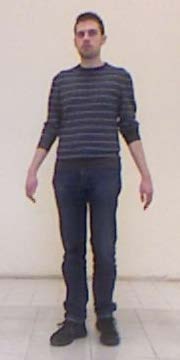}}
           \scalebox{0.11}{\includegraphics{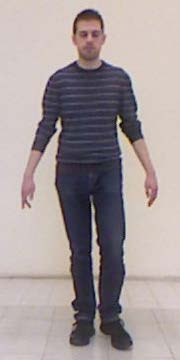}}  
                      \\
      \ \   \scalebox{0.195}{\includegraphics{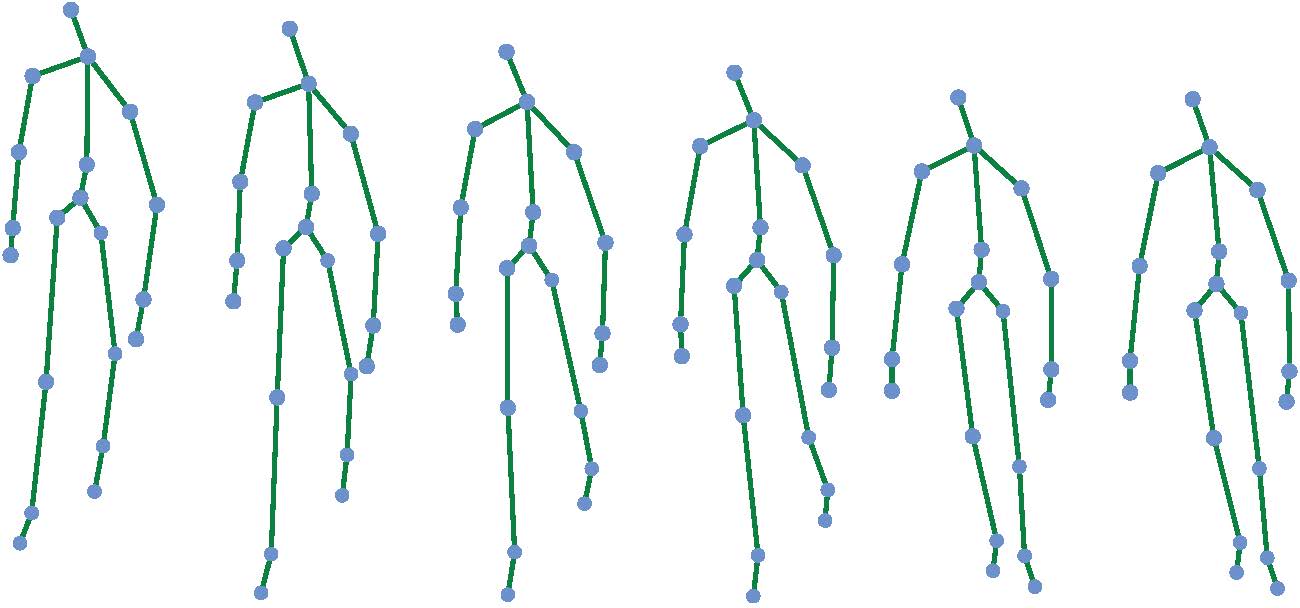}} 
         \ \
          \scalebox{0.085}{\includegraphics{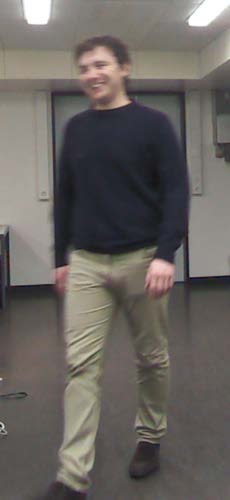}}
       \scalebox{0.085}{\includegraphics{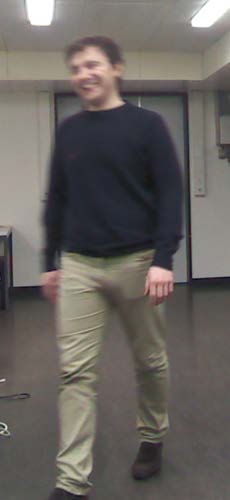}}
        \scalebox{0.085}{\includegraphics{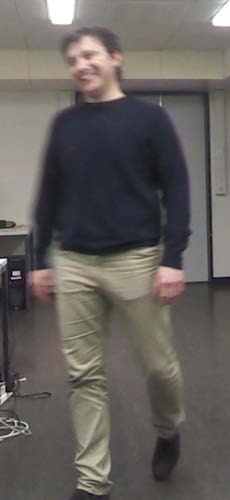}}
         \scalebox{0.085}{\includegraphics{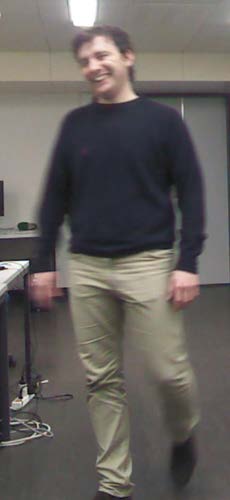}}
          \scalebox{0.085}{\includegraphics{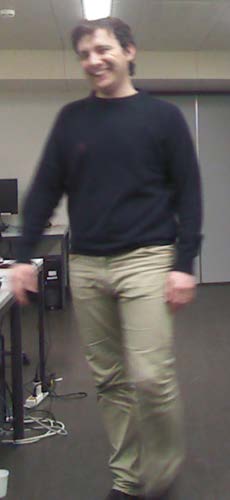}}
           \scalebox{0.085}{\includegraphics{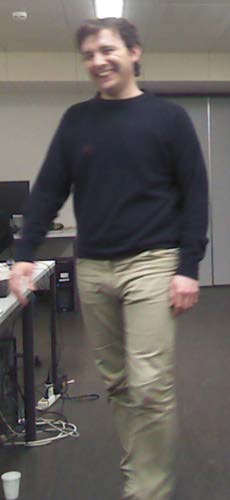}}
\\
            \quad \ \
            \scalebox{0.26}{\includegraphics{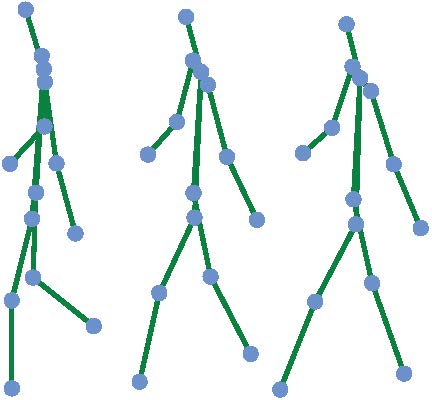}}
         \scalebox{0.26}{\includegraphics{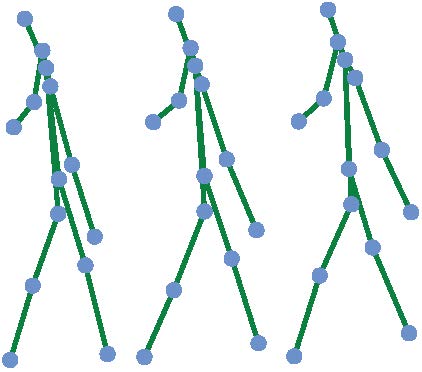}}
     \quad \scalebox{0.067}{\includegraphics{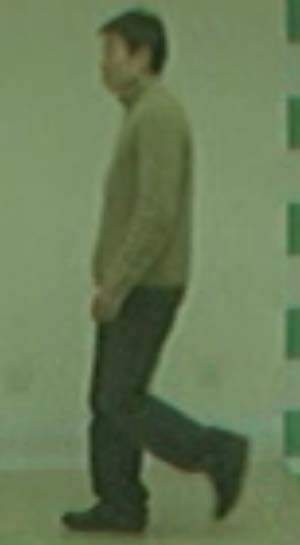}}
       \scalebox{0.067}{\includegraphics{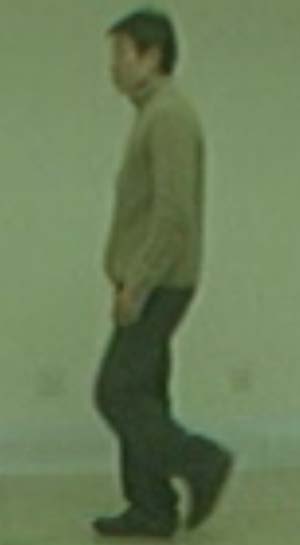}}
        \scalebox{0.067}{\includegraphics{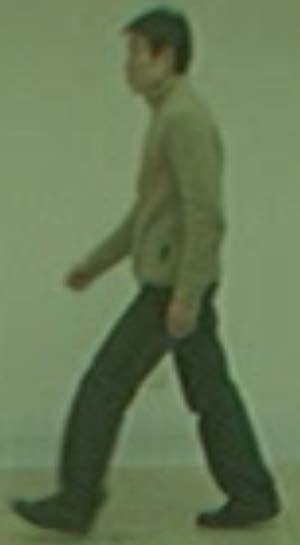}}
         \scalebox{0.067}{\includegraphics{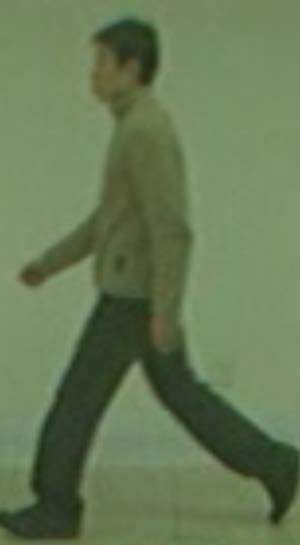}}
          \scalebox{0.067}{\includegraphics{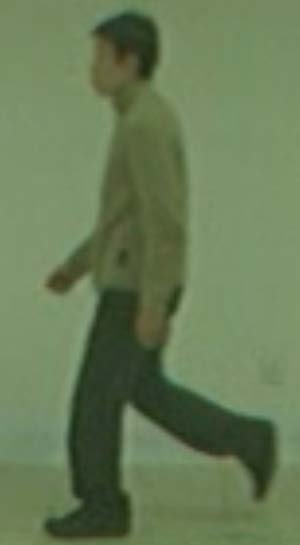}}
           \scalebox{0.067}{\includegraphics{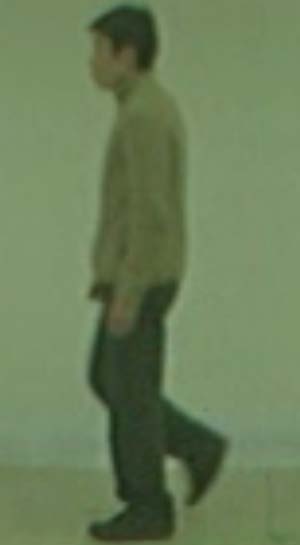}} \quad
    \caption{Examples of 3D skeletons in KGBD (left of $1^{st}$ row), KS20 (right of $1^{st}$ row), IAS ($2^{nd}$ row), BIWI ($3^{rd}$ row), and CASIA-B ($4^{th}$ row). We display both skeleton and RGB samples for RGBD datasets (IAS, BIWI). Note: Skeletons of CASIA-B are estimated from RGB images. }
    \label{sample_skeleton}
\end{figure}

\subsection{Multi-Level Skeleton Meta-Representation}
To combine different-level body and motion semantics embedded in hierarchical skeleton representations and aggregate typical skeleton features learned from different meta-transformed subspaces, we propose to construct \textbf{\textit{multi-level skeleton meta-representation (MSMR)}} as the final skeleton representation with:
\begin{equation}
\scalebox{0.95}{
$
\begin{aligned}
    \boldsymbol{V}&=[\boldsymbol{V}^{1};\boldsymbol{V}^{2};\boldsymbol{V}^{3}]\\
    &=[\frac{1}{M}\sum_{m=1}^{M}(\boldsymbol{\hat{v}}^{1})^{m}; \frac{1}{M}\sum_{m=1}^{M}(\boldsymbol{\hat{v}}^{2})^{m}; \frac{1}{M}\sum_{m=1}^{M}(\boldsymbol{\hat{v}}^{3})^{m}].
    \label{eq_10}
\end{aligned}
$
}
\end{equation}
\hc{In Eq. (\ref{eq_10}),} $\boldsymbol{V}^{l}\in\mathbb{R}^{h}$ denotes the $l^{th}$ level skeleton meta-representation that aggregates features of skeleton meta-instances $(\boldsymbol{\hat{v}}^{l})^{m}$ learned from $M$ meta-transformed subspaces, $[;\!;\!]$ represents feature concatenation, and $\boldsymbol{V}\in\mathbb{R}^{3h}$ is the proposed MSMR that combines different-level skeleton  meta-representations for person re-ID.

\subsection{The Entire Approach}
The computation flow of our approach can be summarized as: $\boldsymbol{S}\rightarrow$
$\boldsymbol{\hat{S}}$ (Sec. \ref{hier_rep_sec})
$\rightarrow\boldsymbol{v}$ (Sec. \ref{Hi-HPC_sec})
$\rightarrow\boldsymbol{\hat{v}}$ $\rightarrow\boldsymbol{\hat{p}}$ 
$\rightarrow\hat{\text{y}}^{m}$ (Sec. \ref{hsm_sec})
$\rightarrow \overline{\delta}((\boldsymbol{\hat{z}}_{j})^{m})$ 
$\rightarrow \mathcal{L}_{\text {Hi-MPC}^{h}}$. 
As illustrated in Algorithm \ref{algorithm_1}, we perform hierarchical skeleton meta-prototype contrastive learning by minimizing $\mathcal{L}_{\text {Hi-MPC}^{h}}$, so as to optimize the encoder $\psi(\cdot)$ and meta-transformation heads to learn effective skeleton representations in an unsupervised manner. During the optimization process, clustering and contrastive learning are alternated to encourage better skeleton representation learning with more reliable clusters.  For the person re-ID task, we encode the probe set $\Phi^{\mathcal{P}}$ into MSMR, $\{\boldsymbol{V}^{\mathcal{P}}_{i}\}_{i=1}^{N_{2}}$, and match it with corresponding representations, $\{\boldsymbol{V}^{\mathcal{G}}_{j}\}_{j=1}^{N_{3}}$, of the same identity in the gallery set $\Phi^{\mathcal{G}}$ using Euclidean distance.

\begin{table}[t]
\caption{Overview of datasets (K: thousand). Different testing splits are used for gallery sets and probe sets (see Sec. \ref{exp_setup}). “W”, “S”, “A”, and “B” denote BIWI-Walking, BIWI-Still, IAS-A, and IAS-B testing sets, respectively. “N”, “C”, and “B” represent “Normal”, “Clothes”, and “Bags” conditions of CASIA-B, respectively. Note: The 3D skeletons of CASIA-B are estimated from RGB videos.}
\label{dataset_details}
\begin{center}
\scalebox{0.735}{
\setlength{\tabcolsep}{2mm}{
\begin{tabular}{l|c|c|c|c|c}
\specialrule{0.1em}{0.45pt}{0.45pt}
                                                                 \textbf{\# Datasets}        & \textbf{KGBD} & \textbf{BIWI}                                                         & \textbf{KS20} & \textbf{IAS}                                                        & \textbf{CASIA-B}                                                               \\ \specialrule{0.1em}{0.45pt}{0.45pt}
\textbf{\# Train IDs}                                                    & 164           & 50                                                                    & 20            & 11                                                                  & 124                                                                            \\ \specialrule{0.1em}{0.45pt}{0.45pt}
\textbf{\begin{tabular}[c]{@{}l@{}}\# Train \\ \ \ \ Skeletons\end{tabular}}  & 188.7K       & 205.8K                                                               & 36.0K        & 89.0K                                                              & 706.5K                                                                        \\ \specialrule{0.1em}{0.45pt}{0.45pt}
\textbf{\# Probe IDs}                                                    & 164           & 28                                                                    & 20            & 11                                                                  & 62                                                                             \\ \specialrule{0.1em}{0.45pt}{0.45pt}
\textbf{\begin{tabular}[c]{@{}l@{}}\# Probe \\ \ \ \ Skeletons\end{tabular}}   & 94.1K        & \begin{tabular}[c]{@{}c@{}}W: 4.9K\\ S: 3.2K\end{tabular} & 3.3K         & \begin{tabular}[c]{@{}c@{}}A: 7.0K\\ B: 7.8K\end{tabular} & \begin{tabular}[c]{@{}c@{}}N: 162.1K\\ C: 54.4K \\ B: 53.9K\end{tabular} \\ \specialrule{0.1em}{0.2pt}{0.2pt}
\textbf{\# Gallery IDs}                                                  & 164           & 28                                                                    & 20            & 11                                                                  & 62                                                                             \\ \specialrule{0.1em}{0.45pt}{0.45pt}
\textbf{\begin{tabular}[c]{@{}l@{}}\# Gallery \\ \ \ \ Skeletons\end{tabular}} & 188.7K       & \begin{tabular}[c]{@{}c@{}}W: 4.9K\\ S: 3.2K\end{tabular} & 3.3K         & \begin{tabular}[c]{@{}c@{}}A: 7.0K\\ B: 7.8K\end{tabular} & \begin{tabular}[c]{@{}c@{}}N: 162.1K\\ C: 54.4K \\ B: 53.9K\end{tabular} \\ \specialrule{0.1em}{0.45pt}{0.45pt}
\end{tabular}
}
}
\end{center}
\end{table}
\section{Experiments}
\label{experiments}
\subsection{Experimental Setup}
\label{exp_setup}
We \rhc{validated} the effectiveness of our approach on four skeleton-based person re-ID benchmarks: \textit{Kinect Gait Biometry Dataset (KGBD)} \cite{andersson2015person}, \textit{BIWI RGBD-ID Dataset (BIWI)} \cite{munaro2014one}, \textit{KS20 VisLab Multi-View Kinect Skeleton Dataset (KS20)} \cite{nambiar2017context}, \textit{IAS-Lab RGBD-ID Dataset (IAS)} \cite{munaro2014feature}, and a large-scale RGB video based multi-view gait dataset: \textit{CASIA-B} \cite{yu2006framework}. \mr{The skeleton data in KGBD, BIWI, KS20, and IAS datasets are captured by the depth sensors like Microsoft Kinect \cite{shotton2011real-time}.} \hc{The original CASIA-B dataset does not contain 3D skeleton data, and we follow \cite{liao2020model} to exploit pre-trained pose estimation models to extract 3D skeletons from RGB videos of CASIA-B, so as to evaluate the performance of our approach on RGB-estimated skeletons.} 
As detailed in \sloppy{Table \ref{dataset_details}}, these five datasets contain skeleton data of 164, 50, 20, 11, and 124 different identities, respectively.

We follow the commonly-used settings of probe and gallery in the literature \cite{rao2022simmc}:  For \rhc{the} BIWI and IAS datasets, as different testing sets are non-overlapped and contain all pedestrians under different scenes, we evaluate our approach on each testing set by setting it as the probe while the other one is adopted as the gallery. The KGBD dataset contains different skeleton videos ($i.e.,$ long skeleton sequences) of each pedestrian with varying numbers of walking rounds. Since no training/testing splits are given, we randomly \rhc{choose one skeleton video of each person to split skeleton sequences and construct the probe set, and equally divide the remaining videos to build the training set and gallery set.}
 The KS20 dataset collects skeleton data of pedestrians from five different viewpoints, including $0^{\circ}$, $30^{\circ}$, $90^{\circ}$, $130^{\circ}$, and $180^{\circ}$.
 We employ different splitting setups to evaluate the multi-view person re-ID performance of our approach. For Random View Evaluation (RVE), one sequence is randomly selected from each viewpoint as the probe sequence and the remaining skeleton sequences are equally divided into gallery and training sequences. In Cross-View Evaluation (CVE) setup, we match persons across views, $i.e.,$ matching between two different views that are not involved in training. We set sequences from two different viewpoints as the probe set and gallery set, respectively, while adopting sequences from \rhc{the} remaining viewpoints as the training set. The CASIA-B dataset contains sequences of 124 individuals under 11 different views and 3 conditions---pedestrians wearing a bag (``Bags''), wearing a coat (``Clothes''), and without any coat or bag (``Normal''). We follow \rhc{the} person re-ID protocols in \cite{liu2015enhancing} to evaluate the proposed skeleton-based approach on CASIA-B.
 Experiments with each setup are repeated for multiple times and the average performance is reported in this work. More details about the experimental setups are given in Appendix \uppercase\expandafter{\romannumeral2}.

\subsection{Implementation Details} 
The numbers of body joints in skeletons are $J=25$ for KS20, $J=20$ for KGBD, BIWI, IAS, and $J=14$ for CASIA-B. 
For fair comparison with existing methods, we set the sequence length $F=6$ on four skeleton-based datasets (KS20, KGBD, BIWI, IAS) and $F=40$ on estimated skeleton data of CASIA-B.
\mr{
Three level ($i.e.,$ joint-level, component-level, limb-level) hierarchical skeleton representations are combined as the final configuration to construct MSMR in the experiments.}
The embedding feature size of skeleton representations is $h=256$, and the number of meta-transformation heads is $M=8$ for all datasets. We empirically set \hc{the} maximum distance \hc{in the} DBSCAN algorithm to $0.6$ (KGBD, BIWI-S), $0.8$ (KS20, IAS, BIWI-W), $0.75$ (CASIA-B), and employ minimum amount of samples with $4$ for KGBD and $2$ for other datasets.
The Adam optimizer is used with learning rate $0.00035$ and the batch size is set to $256$ on all datasets. 
We present full implementation details in the appendices and our source code is released at  \href{https://github.com/Kali-Hac/Hi-MPC}{https://github.com/Kali-Hac/Hi-MPC}.

\subsection{Evaluation Metrics} 
We compute \rhc{the} Cumulative Matching Characteristics (CMC) curve and adopt Rank-1 accuracy (R$_{1}$), Rank-5 accuracy (R$_{5}$), and Rank-10 accuracy (R$_{10}$) as performance metrics \cite{rao2022simmc}. R$_{1}$, R$_{5}$, and R$_{10}$ are computed as the ratios of probe sequences matching the gallery sequences with correct identities when the candidate gallery sequences are the top 1, top 1 to 5, and top 1 to 10 most similar sequences to the probe sequence. Mean Average Precision (mAP) \cite{zheng2015scalable} is also used to quantitatively evaluate the overall performance of our approach. 

\begin{table*}[t]
\centering
\caption{Performance comparison with state-of-the-art skeleton-based methods on BIWI-S, BIWI-W, and IAS-A. ${\ddagger}$ indicates employing supervised fine-tuning and \textbf{Bold} denotes the best cases among self-supervised/unsupervised methods. \mr{The $\underline{\text{underline}}$ represents the highest results among all methods.}}
\label{Results-1}
\scalebox{0.7}{
\setlength{\tabcolsep}{2.8mm}{
\begin{tabular}{llrrrrrrrrrrrr}
\hline
\textbf{}                                                                                         & \textbf{}                                         & \multicolumn{4}{c}{\textbf{BIWI-S}}                                                                                                                                                     & \multicolumn{4}{c}{\textbf{BIWI-W}}                                                                                                                                                     & \multicolumn{4}{c}{\textbf{IAS-A}}                                                                                                                                                      \\ \hline
\textbf{Types}                                                                                    & \textbf{Methods}                                   & \textbf{R$_{1}$} & \textbf{R$_{5}$} & \textbf{R$_{10}$} & \textbf{mAP}  & \textbf{R$_{1}$} & \textbf{R$_{5}$} & \textbf{R$_{10}$} & \textbf{mAP}  & \textbf{R$_{1}$} & \textbf{R$_{5}$} & \textbf{R$_{10}$} & \textbf{mAP}  \\ \hline
\multirow{3}{*}{\textbf{Hand-crafted}}      
& \mr{$D_{\text{PG}}$ \cite{liao2020model}} & 18.5             & 45.4             & 63.8              & 6.7           & 6.5              & 15.5             & 20.3              & 8.7           & 16.4             & 39.5             & 53.4              & 11.0          \\
& \mr{$D_{13}$ \cite{munaro2014one}}    & 28.3                                            & 53.1                                            & 65.9                                             & 13.1                             & 14.2                                            & 20.6                                            & 23.7                                             & 17.2                             & 40.0                                            & 58.7                                            & 67.6                                             & 24.5                             \\
              & \mr{$D_{16}$ \cite{pala2019enhanced}} & 32.6                                            & 55.7                                            & 68.3                                             & 16.7                             & 17.0                                            & 25.3                                            & 29.6                                             & 18.8                             & 42.7                                            & 62.9                                            & 70.7                                             & \underline{25.2}                             \\ \hline
\multirow{4}{*}{\textbf{Supervised}}                                                              & PoseGait \cite{liao2020model}    & 14.0                                            & 40.7                                            & 56.7                                             & 9.9                              & 8.8                                             & 23.0                                            & 31.2                                             & 11.1                             & 28.4                                            & 55.7                                            & 69.2                                             & 17.5                             \\
              & $^{\ddagger}$SGELA \cite{rao2021self}     & 29.2                                            & 65.2                                            & 73.8                                             & \underline{23.5}                             & 13.9                                            & 15.3                                            & 16.7                                             & \underline{22.9}                             & 18.0                                            & 32.1                                            & 46.2                                             & 13.5                             \\
              & MG-SCR \cite{rao2021multi}       & 20.1                                            & 46.9                                            & 64.1                                             & 7.6                              & 10.8                                            & 20.3                                            & 29.4                                             & 11.9                             & 36.4                                            & 59.6                                            & 69.5                                             & 14.1                             \\
              & $^{\ddagger}$SM-SGE \cite{rao2021sm}     & 34.8                                            & 60.6                                            & 71.5                                             & 12.8                             & 16.7                                            & 31.0                                            & 40.2                                             & 18.7                             & 38.5                                            & 63.2                                            & 73.9                                             & 15.0                             \\ \hline
\multirow{5}{*}{\textbf{\begin{tabular}[c]{@{}l@{}}Self-supervised\\ /Unsupervised\end{tabular}}} & AGE \cite{rao2020self}           & 25.1                                            & 43.1                                            & 61.6                                             & 8.9                              & 11.7                                            & 21.4                                            & 27.3                                             & 12.6                             & 31.1                                            & 54.8                                            & 67.4                                             & 13.4                             \\
              & SGELA \cite{rao2021self}         & 25.8                                            & 51.8                                            & 64.4                                             & 15.1                             & 11.7                                            & 14.0                                            & 14.7                                             & 19.0                             & 16.7                                            & 30.2                                            & 44.0                                             & 13.2                             \\
              & SM-SGE \cite{rao2021sm}          & 31.3                                            & 56.3                                            & 69.1                                             & 10.1                             & 13.2                                            & 25.8                                            & 33.5                                             & 15.2                             & 34.0                                            & 60.5                                           & 71.6                                             & 13.6                             \\
              & SimMC \cite{rao2022simmc}                                            & 41.7                                            & 66.6                                            & 76.8                                             & 12.3                             & 24.5                                            & 36.7                                            & 44.5                                             & 19.9                             & 44.8                                            & 65.3                                            & 72.9                                             & 18.7                             \\
              & \textbf{Hi-MPC$^{h}$ (Ours)}                      & \underline{\textbf{47.5}}                                   & \underline{\textbf{70.3}}                                   & \underline{\textbf{78.6}}                                    & \textbf{17.4}                    & \underline{\textbf{27.3}}                                   & \underline{\textbf{40.3}}                                   & \underline{\textbf{48.8}}                                    & \textbf{22.6}                    & \underline{\textbf{45.6}}                                   & \underline{\textbf{67.3}}                                   & \underline{\textbf{75.4}}                                    & \textbf{23.2}                    \\ \hline
\end{tabular}
}
}
\end{table*}

\begin{table*}[t]
\centering
\caption{Performance comparison with state-of-the-art skeleton-based methods on IAS-B, KGBD, and KS20. \mr{${\ddagger}$ indicates employing supervised fine-tuning and \textbf{Bold} denotes the best cases among self-supervised/unsupervised methods. The $\underline{\text{underline}}$ represents the highest results among all methods. }}
\label{Results-2}
\scalebox{0.7}{
\setlength{\tabcolsep}{2.95mm}{
\begin{tabular}{llrrrrrrrrrrrr}
\hline
\textbf{}                                                                                         & \textbf{}                                         & \multicolumn{4}{c}{\textbf{IAS-B}}                                                                                                                                                      & \multicolumn{4}{c}{\textbf{KGBD}}                                                                                                                                                       & \multicolumn{4}{c}{\textbf{KS20}}                                                                                                                                                       \\ \hline
\textbf{Types}                                                                                    & \textbf{Methods}                                 & \textbf{R$_{1}$} & \textbf{R$_{5}$} & \textbf{R$_{10}$} & \textbf{mAP}  & \textbf{R$_{1}$} & \textbf{R$_{5}$} & \textbf{R$_{10}$} & \textbf{mAP}  & \textbf{R$_{1}$} & \textbf{R$_{5}$} & \textbf{R$_{10}$} & \textbf{mAP}  \\ \hline
\multirow{3}{*}{\textbf{Hand-crafted}}  
 & \mr{$D_{\text{PG}}$ \cite{liao2020model}} & 16.0             & 41.2             & 57.3              & 10.6          & 30.0             & 49.1             & 58.1              & 2.1           & 35.2             & 61.5             & 70.5              & 11.3          \\ & \mr{$D_{13}$ \cite{munaro2014one}}    & 43.7                                            & 68.6                                            & 76.7                                             & 23.7                             & 17.0                                            & 34.4                                            & 44.2                                             & 1.9                              & 39.4                                            & 71.7                                            & 81.7                                             & 18.9                             \\
          & \mr{$D_{16}$ \cite{pala2019enhanced}} & 44.5                                            & 69.1                                            & \underline{80.2}                                             & 24.5                             & 31.2                                            & 50.9                                            & 59.8                                             & 4.0                              & 51.7                                            & 77.1                                            & 86.9                                             & \underline{24.0}                             \\ \hline
\multirow{4}{*}{\textbf{Supervised}}                                                              & PoseGait \cite{liao2020model}    & 28.9                                            & 51.6                                            & 62.9                                             & 20.8                             & 50.6                                            & 67.0                                            & 72.6                                             & \underline{13.9}                             & 49.4                                            & 80.9                                            & \underline{90.2}                                             & 23.5                             \\
          & $^{\ddagger}$SGELA \cite{rao2021self}     & 23.6                                            & 42.9                                            & 51.9                                             & 14.8                             & 43.7                                            & 58.7                                            & 65.0                                             & 7.1                              & 49.7                                            & 67.0                                            & 77.1                                             & 22.2                             \\
          & MG-SCR \cite{rao2021multi}       & 32.4                                            & 56.5                                            & 69.4                                             & 12.9                             & 44.0                                            & 58.7                                            & 64.6                                             & 6.9                              & 46.3                                            & 75.4                                            & 84.0                                             & 10.4                             \\
          & $^{\ddagger}$SM-SGE \cite{rao2021sm}      & 44.3                                            & 68.2                                            & 77.5                                             & 14.9                             & 43.2                                            & 58.6                                            & 64.6                                             & 7.5                              & 49.8                                            & 78.1                                            & 85.2                                             & 11.7                             \\ \hline
\multirow{5}{*}{\textbf{\begin{tabular}[c]{@{}l@{}}Self-supervised\\ /Unsupervised\end{tabular}}} & AGE \cite{rao2020self}           & 31.1                                            & 52.3                                            & 64.2                                             & 12.8                             & 2.9                                             & 5.6                                             & 7.5                                              & 0.9                              & 43.2                                            & 70.1                                            & 80.0                                             & 8.9                              \\
          & SGELA \cite{rao2021self}         & 22.2                                            & 40.8                                            & 50.2                                             & 14.0                             & 38.1                                            & 53.5                                            & 60.0                                             & 4.5                              & 45.0                                            & 65.0                                            & 75.1                                             & 21.2                             \\
          & SM-SGE \cite{rao2021sm}          & 38.9                                            & 64.1                                            & 75.8                                             & 13.3                             & 38.2                                            & 54.2                                          & 60.7                                            & 4.4                              & 45.9                                            & 71.9                                            & 81.2                                             & 9.5                              \\
          & SimMC \cite{rao2022simmc}                                            & 46.3                                            & 68.1                                            & 77.0                                             & 22.9                             & 54.9                                            & 66.2                                            & 70.6                                             & \textbf{11.7}                    & 66.4                                            & 80.7                                            & 87.0                                             & \textbf{22.3}                    \\
          & Hi-MPC$^{h}$ (Ours)                               & \underline{\textbf{48.2}}                                            & \underline{\textbf{70.2}}                                            & \textbf{77.8}                                    & \underline{\textbf{25.3}}                             & \underline{\textbf{56.9}}                                   & \underline{\textbf{70.2}}                                   & \underline{\textbf{75.1}}                                    & 10.2                             & \underline{\textbf{69.6}}                                   & \underline{\textbf{83.5}}                                   & \textbf{87.1}                                    & 22.0                             \\ \hline
\end{tabular}
}
}
\end{table*}

\begin{table*}[t]
\centering
\caption{Cross-view person re-ID performance comparison with state-of-the-art self-supervised and unsupervised methods with CVE setup of KS20. $0^{\circ}, 30^{\circ}, 90^{\circ}, 130^{\circ}$, and $180^{\circ}$ denote different views of probe or gallery sets.}
\label{KS20_CVE}
\scalebox{0.68}{
\setlength{\tabcolsep}{0.7mm}{
\begin{tabular}{clrrrrrrrrrrrrrrrrrrrr}
\hline
\multicolumn{1}{l}{}                     & \textbf{Gallery Views}       & \multicolumn{4}{c}{\textbf{$0^{\circ}$}}                                & \multicolumn{4}{c}{\textbf{$30^{\circ}$}}                               & \multicolumn{4}{c}{\textbf{$90^{\circ}$}}                               & \multicolumn{4}{c}{\textbf{$130^{\circ}$}}                              & \multicolumn{4}{c}{\textbf{$180^{\circ}$}}                              \\ \hline
\multicolumn{1}{l}{\textbf{Probe Views}} & \textbf{}                    & \textbf{R$_{1}$} & \textbf{R$_{5}$} & \textbf{R$_{10}$} & \textbf{mAP}  & \textbf{R$_{1}$} & \textbf{R$_{5}$} & \textbf{R$_{10}$} & \textbf{mAP}  & \textbf{R$_{1}$} & \textbf{R$_{5}$} & \textbf{R$_{10}$} & \textbf{mAP}  & \textbf{R$_{1}$} & \textbf{R$_{5}$} & \textbf{R$_{10}$} & \textbf{mAP}  & \textbf{R$_{1}$} & \textbf{R$_{5}$} & \textbf{R$_{10}$} & \textbf{mAP}  \\ \hline
\multirow{5}{*}{\textbf{$0^{\circ}$}}    & AGE \cite{rao2020self}                          & 46.7             & 74.2             & 83.5              & 22.5          & 11.0             & 35.7             & 47.5              & 10.0          & 8.1              & 29.9             & 47.5              & 9.2           & 7.5              & 26.7             & 43.5              & 8.4           & 7.0              & 23.0             & 37.4              & 8.2           \\
                                         & SGELA \cite{rao2021self}                        & 76.2             & 89.6             & 92.8              & 37.1          & 15.1             & 27.3             & 35.1              & 19.9          & 10.1             & 27.5             & 40.9              & 18.2          & 10.7             & 21.5             & 29.3              & 18.0          & 15.4             & 25.8             & 38.0              & 12.6          \\
                                         & SM-SGE \cite{rao2021sm}                       & 58.4             & 84.7             & 92.2              & 27.7          & 17.2             & 50.0             & 63.3              & 10.8          & 7.2              & 21.9             & 39.1              & 10.5          & 4.4              & 19.4             & 34.7              & 9.3           & 10.0             & 23.8             & 33.1              & 9.4           \\
                                         & SimMC \cite{rao2022simmc}                       & 84.4             & 97.3             & 99.2              & 61.2          & \textbf{37.9}    & 59.4             & 67.6              & 24.8          & 28.9             & 50.8             & 62.9              & \textbf{27.1} & 23.3             & \textbf{43.0}    & 52.9              & \textbf{20.3} & 15.2             & 29.3             & \textbf{45.7}     & \textbf{14.5} \\
                                         & Hi-MPC$^{h}$ (Ours)          & \textbf{92.2}    & \textbf{98.4}    & \textbf{99.6}     & \textbf{66.9} & 35.6             & \textbf{61.7}    & \textbf{72.7}     & \textbf{27.3} & \textbf{36.7}    & \textbf{58.6}    & \textbf{73.1}     & 25.9          & \textbf{24.6}    & 41.8             & \textbf{53.5}     & 16.4          & \textbf{17.2}    & \textbf{30.1}    & 43.8              & 13.2          \\ \hline
\multirow{5}{*}{\textbf{$30^{\circ}$}}   & AGE                          & 10.1             & 42.8             & 57.8              & 8.8           & 52.3             & 82.7             & 91.5              & 25.0          & 15.0             & 35.6             & 58.5              & 8.8           & 10.1             & 24.2             & 41.8              & 8.1           & 7.8              & 24.2             & 34.3              & 8.3           \\
                                         & SGELA                        & 13.1             & 19.6             & 22.6              & 19.4          & 70.9             & 88.2             & 91.8              & 40.5          & 11.8             & 24.5             & 36.3              & 16.5          & 6.9              & 22.6             & 31.7              & 15.4          & 9.2              & 15.4             & 22.9              & 13.9          \\
                                         & SM-SGE                       & 18.1             & 48.4             & 65.0              & 11.5          & 60.2             & 82.0             & 89.8              & 28.2          & 12.5             & 27.2             & 35.3              & 10.7          & 7.5              & 23.4             & 33.8              & 10.6          & 8.8              & 27.2             & 39.1              & 10.5          \\
                                         & SimMC                        & 30.8             & 66.2             & 74.6              & 20.7          & 91.8             & 97.4             & 98.2              & \textbf{67.8} & 36.8             & 55.1             & 67.6              & \textbf{29.9} & 16.4             & 30.5             & 40.2              & \textbf{20.4} & 16.2             & 36.7             & \textbf{56.6}     & 12.7          \\
                                         & Hi-MPC$^{h}$ (Ours)          & \textbf{33.2}    & \textbf{66.4}    & \textbf{75.8}     & \textbf{24.4} & \textbf{93.8}    & \textbf{97.7}    & \textbf{98.8}     & 66.5          & \textbf{37.5}    & \textbf{62.1}    & \textbf{71.5}     & 25.1          & \textbf{19.5}    & \textbf{34.8}    & \textbf{50.0}     & 17.8          & \textbf{16.8}    & \textbf{37.5}    & 47.7              & \textbf{14.3} \\ \hline
\multirow{5}{*}{\textbf{$90^{\circ}$}}   & AGE                          & 7.5              & 27.3             & 43.2              & 8.7           & 9.0              & 28.5             & 44.1              & 9.3           & 57.4             & 81.4             & 90.7              & 19.2          & 13.8             & 41.1             & 57.1              & 9.0           & 7.8              & 30.0             & 46.0              & 8.3           \\
                                         & SGELA                        & 9.6              & 19.8             & 29.7              & 16.4          & 10.8             & 15.6             & 20.4              & 17.5          & 48.4             & 75.7             & 86.5              & 31.6          & 17.1             & 35.7             & 43.0              & 22.0          & 13.5             & 23.4             & 31.8              & 21.3          \\
                                         & SM-SGE                       & 19.1             & 33.1             & 48.1              & 12.4          & 23.1             & 40.6             & 57.4              & 11.5          & 72.2             & 89.1             & 92.8              & 24.9          & 20.9             & 48.4             & 69.4              & 12.8          & 19.4             & 36.9             & 51.6              & 11.3          \\
                                         & SimMC                        & \textbf{26.2}    & 44.9             & 50.8              & 11.9          & 41.4             & 64.1             & 75.4              & 27.3          & 96.7             & \textbf{100}     & \textbf{100}      & 73.1          & 60.9             & \textbf{81.6}    & \textbf{88.7}     & \textbf{45.0} & 25.8             & 48.4             & 64.5              & 15.4          \\
                                         & Hi-MPC$^{h}$ (Ours)          & \textbf{26.2}    & \textbf{47.7}    & \textbf{62.1}     & \textbf{23.0} & \textbf{50.8}    & \textbf{71.5}    & \textbf{83.2}     & \textbf{34.0} & \textbf{97.3}    & \textbf{100}     & \textbf{100}      & \textbf{73.9} & \textbf{61.7}    & 80.5             & 84.8              & 42.2          & \textbf{33.2}    & \textbf{65.2}    & \textbf{80.1}     & \textbf{23.1} \\ \hline
\multirow{5}{*}{\textbf{$130^{\circ}$}}  & AGE                          & 6.7              & 21.3             & 34.7              & 8.2           & 7.9              & 23.4             & 38.9              & 8.9           & 15.2             & 35.9             & 54.4              & 9.2           & 45.3             & 70.5             & 82.1              & 18.7          & 11.3             & 37.1             & 50.2              & 8.9           \\
                                         & SGELA                        & 5.8              & 18.8             & 28.0              & 14.2          & 11.6             & 15.5             & 20.7              & 16.8          & 17.6             & 47.1             & 53.2              & 24.5          & 59.6             & 81.5             & 89.1              & 36.8          & 17.0             & 29.8             & 32.5              & 23.0          \\
                                         & SM-SGE                       & 8.4              & 24.4             & 37.8              & 10.4          & 12.9             & 26.6             & 36.3              & 10.9          & 24.1             & 53.4             & 66.3              & 12.9          & 64.4             & 85.9             & 95.0              & 25.5          & 17.8             & 40.9             & 59.1              & 12.1          \\
                                         & SimMC                        & 18.0             & 32.4             & 48.8              & 14.2          & \textbf{24.2}    & 44.9             & 59.4              & 15.7          & 60.2             & 78.1             & 86.7              & 45.2          & 92.5             & 98.8             & \textbf{99.2}     & 71.5          & 30.1             & 55.1             & 66.8              & 18.8          \\
                                         & Hi-MPC$^{h}$ (Ours)          & \textbf{19.9}    & \textbf{39.1}    & \textbf{55.1}     & \textbf{20.3} & 20.7             & \textbf{50.8}    & \textbf{68.4}     & \textbf{21.9} & \textbf{62.1}    & \textbf{80.5}    & \textbf{87.5}     & \textbf{45.8} & \textbf{93.4}    & \textbf{99.2}    & \textbf{99.2}     & \textbf{72.8} & \textbf{36.3}    & \textbf{61.3}    & \textbf{75.8}     & \textbf{26.3} \\ \hline
\multirow{5}{*}{\textbf{$180^{\circ}$}}  & AGE                          & 7.9              & 17.7             & 32.6              & 8.1           & 5.2              & 22.4             & 33.4              & 8.3           & 10.5             & 25.6             & 34.0              & 8.2           & 11.6             & 33.1             & 52.9              & 8.8           & 47.1             & 72.4             & 82.6              & 22.6          \\
                                         & SGELA                        & 14.0             & 29.1             & 39.2              & 21.3          & 11.9             & 20.6             & 25.9              & \textbf{17.3} & 18.6             & 37.8             & 49.7              & 19.4          & 22.7             & 45.9             & 55.2              & 20.7          & 74.5             & 92.7             & 95.1              & 38.3          \\
                                         & SM-SGE                       & 5.6              & 20.0             & 30.6              & 8.5           & 6.6              & 22.7             & 31.6              & 8.6           & 13.8             & 34.1             & 45.6              & 9.4           & 10.3             & 37.5             & 56.6              & 10.4          & 51.9             & 79.7             & 87.8              & 25.6          \\
                                         & SimMC                        & \textbf{19.1}    & \textbf{39.5}    & \textbf{48.8}     & 15.3          & 14.1             & 28.1             & 40.6              & 14.0          & 25.8             & 44.1             & 53.5              & \textbf{29.8} & 36.9             & 61.3             & 74.6              & \textbf{32.0} & 91.0             & \textbf{97.7}    & 98.4              & \textbf{59.5} \\
                                         & Hi-MPC$^{h}$ (Ours) & 16.8             & 34.4             & 47.6              & \textbf{15.6} & \textbf{14.8}    & \textbf{30.9}    & \textbf{44.5}     & 14.8          & \textbf{30.9}    & \textbf{55.5}    & \textbf{68.0}     & 25.8          & \textbf{37.1}    & \textbf{62.5}    & \textbf{79.3}     & 29.5          & \textbf{93.0}    & 97.3             & \textbf{98.8}     & 58.0          \\ \hline
\end{tabular}
}
}
\end{table*}

\subsection{Empirical Evaluation}
\label{performance_comparison}
\mr{We compare our \textit{unsupervised} approach with state-of-the-art unsupervised skeleton-based person re-ID methods on BIWI, IAS, KGBD, and KS20 datasets, as shown in \sloppy{Tables \ref{Results-1} and \ref{Results-2}}. 
The latest self-supervised, supervised skeleton-based person re-ID methods \cite{liao2020model,rao2021multi,rao2021sm} and representative hand-crafted person re-ID methods \cite{munaro2014one,pala2019enhanced} are also included to provide a comprehensive comparison. It should be noted that the hand-crafted methods directly extract skeleton descriptors from raw skeleton data to perform person re-ID.}

\subsubsection{Comparison with Self-Supervised and Unsupervised \rhc{State-of-the-Art Methods}}
The proposed approach shows significant performance \hc{improvements} over existing self-supervised and unsupervised methods on different datasets. Compared with AGE \cite{rao2020self} and SGELA \cite{rao2021self} that learn skeleton features from only joint-level representations, our approach achieves higher person re-ID performance on all datasets with an improvement of $14.5$-$54.0\%$ \hc{for} Rank-1 accuracy, $12.5$-$64.6\%$ \hc{for} Rank-5 accuracy, $7.1$-$67.6\%$ \hc{for} Rank-10 accuracy, and $0.8$-$13.1\%$ \hc{for} mAP. \hc{These improvements indicate} that the proposed approach can capture more discriminative features from multiple skeleton levels for person re-ID. By mining key skeletons and exploiting more informative patterns of different levels with the proposed HSM mechanism, our approach \hc{significantly} outperforms the SM-SGE framework \cite{rao2021sm} that directly combines four-scale skeletons by \hc{$9.3$-$23.7\%$ for Rank-1 accuracy and $5.8$-$12.5\%$ for mAP} on different datasets. 
\hc{Note that we follow \cite{rao2021self,rao2022simmc} to report the average performance of all methods for a fair comparison while there exist slight performance variations in practice. In this context, the proposed approach is still superior to the latest skeleton contrastive learning framework SimMC \cite{rao2022simmc} in most cases, as shown in \sloppy{\rhc{Tables} \ref{Results-1} and \ref{Results-2}}. Although SimMC obtains slightly higher mAP on KS20 and KGBD, our approach can achieve better overall performance in terms of Rank-1, Rank-5, Rank-10, and mAP on the datasets that contain \rhc{frequent changes of appearances or scenes} (BIWI-S, BIWI-W, IAS-A, IAS-B). Considering that \rhc{these} changes could induce more random perturbations or noises in the collection of skeletons, \rhc{the results suggest that our approach is more robust than} SimMC for learning effective skeleton representations under different conditions. \rhc{Moreover}, our approach is more efficient than most skeleton-based person re-ID methods in terms of the model size and computational complexity (see Appendix \uppercase\expandafter{\romannumeral2}).}

%
Our approach is also evaluated on the cross-view person re-ID scenarios of KS20. As presented in \sloppy{Table \ref{KS20_CVE}}, the proposed Hi-MPC$^{h}$ outperforms state-of-the-art self-supervised and unsupervised counterparts by an average margin of $6.4$ to $44.4\%$ \hc{for} Rank-1 accuracy and $3.0$ to $48.7\%$ \hc{for} mAP on all views, and also surpasses the latest SimMC framework \cite{rao2022simmc} in most probe-gallery matching views. \hc{This} suggests that our model is more effective on learning generalized ($i.e.,$ view-independent) skeleton representations with higher robustness to viewpoint changes for cross-view person re-ID.


\subsubsection{Comparison with Hand-Crafted and Supervised \rhc{State-of-the-Art Methods}}
The results in \sloppy{\rhc{Tables} \ref{Results-1} and \ref{Results-2}} show that our model achieves better person re-ID performance than the representative hand-crafted methods $D_{13}$ \cite{munaro2014one} and $D_{16}$ \cite{pala2019enhanced} that utilize numerous anthropometric skeleton descriptors by $3.7$-$39.9\%$ \hc{for} Rank-1 accuracy and $0.7$-$8.3\%$ \hc{for} mAP on \hc{four of the six} testing sets (BIWI-S, BIWI-W, IAS-B, KGBD). Although they achieve competitive mAP on datasets with frequent appearance and viewpoint changes (IAS-A, KS20), our approach can obtain higher overall performance in terms of Rank-1 accuracy ($2.9$-$30.2\%$), Rank-5 accuracy ($4.4$-$11.8\%$), and Rank-10 accuracy ($0.2$-$7.8\%$). Notably, the proposed unsupervised approach markedly outperforms supervised state-of-the-art models PoseGait \cite{liao2020model} and MG-SCR \cite{rao2021multi} on almost all datasets. Interestingly, with extra labels to fine-tune SGELA \cite{rao2021self} and SM-SGE \cite{rao2021sm}, those methods still \hc{perform poorly} on many datasets. In contrast, our approach achieves \hc{better and more stable} performance on all datasets without using any skeletal annotation. \mr{This shows that Hi-MPC$^{h}$ has good generality with greater potential for use in practical person re-ID scenarios.}


\section{Further Analysis}
\label{discussion}

\subsection{Evaluation on \mr{RGB-Estimated} Skeleton Data}
\begin{table*}[t]
\centering
\caption{Performance comparison with appearance-based and skeleton-based methods on CASIA-B. ``Clothes-Normal'' represents the probe set under ``Clothes'' condition and gallery set under ``Normal'' condition. ${\clubsuit}$ refers to appearance-based methods and ${\ddagger}$ represents requiring label information for training. ``—'' indicates no published result.}
\label{CASIA_B_result}
\scalebox{0.7}{
\setlength{\tabcolsep}{0.75mm}{
\begin{tabular}{lrrrrrrrrrrrrrrrrrrrr}
\hline
\textbf{Probe-Gallery}                                     & \multicolumn{4}{c}{\textbf{Normal-Normal}}                                                                                                                                                      & \multicolumn{4}{c}{\textbf{Bags-Bags}}                                                                                                                                                      & \multicolumn{4}{c}{\textbf{Clothes-Clothes}}                                                                                                                                                      & \multicolumn{4}{c}{\textbf{Clothes-Normal}}                                                                                                                                                      & \multicolumn{4}{c}{\textbf{Bags-Normal}}                                                                                                               \\ \hline
\textbf{Methods}                                           & \textbf{R$_{1}$} & \textbf{R$_{5}$} & \textbf{R$_{10}$} & \textbf{mAP}  & \textbf{R$_{1}$} & \textbf{R$_{5}$} & \textbf{R$_{10}$} & \textbf{mAP}  & \textbf{R$_{1}$} & \textbf{R$_{5}$} & \textbf{R$_{10}$} & \textbf{mAP}  & \textbf{R$_{1}$} & \textbf{R$_{5}$} & \textbf{R$_{10}$} & \textbf{mAP} & \textbf{R$_{1}$} & \textbf{R$_{5}$} & \textbf{R$_{10}$} & \textbf{mAP} \\ \hline
$^{\ddagger}$LMNN$^{\clubsuit}$ \cite{weinberger2009distance}        & 3.9                                             & 22.7                                            & 36.1                                             & —                                & 18.3                                            & 38.6                                            & 49.2                                             & —                                & 17.4                                            & 35.7                                            & 45.8                                             & —                                & 11.6                                            & 12.6                                            & 17.8                                             & —                                & 23.1                               & 37.1                               & 44.4                                & —                                \\
$^{\ddagger}$ITML$^{\clubsuit}$ \cite{davis2007information}          & 7.5                                             & 22.2                                            & 34.2                                             & —                                & 19.5                                            & 26.0                                            & 33.7                                             & —                                & 20.1                                            & 34.4                                            & 43.3                                             & —                                & 10.3                                            & 24.5                                            & 36.1                                             & —                                & 21.8                               & 30.4                               & 36.3                                & —                                \\
$^{\ddagger}$ELF$^{\clubsuit}$ \cite{gray2008viewpoint}              & 12.3                                            & 35.6                                            & 50.3                                             & —                                & 5.8                                             & 25.5                                            & 37.6                                             & —                                & 19.9                                            & 43.9                                            & 56.7                                             & —                                & 5.6                                             & 16.0                                            & 26.3                                             & —                                & 17.1                               & 30.0                               & 37.9                                & —                                \\
SDALF$^{\clubsuit}$ \cite{farenzena2010person}          & 4.9                                             & 27.0                                            & 41.6                                             & —                                & 10.2                                            & 33.5                                            & 47.2                                             & —                                & 16.7                                            & 42.0                                            & 56.7                                             & —                                & 11.6                                            & 19.4                                            & 27.6                                             & —                                & 22.9                               & 30.1                               & 36.1                                & —                                \\
$^{\ddagger}$Score-based MLR$^{\clubsuit}$ \cite{liu2015enhancing}   & 13.6                                            & 48.7                                            & 63.7                                             & —                                & 13.6                                            & 48.7                                            & 63.7                                             & —                                & 13.5                                            & 48.6                                            & 63.9                                             & —                                & 9.7                                             & 27.8                                            & 45.1                                             & —                                & 14.7                               & 32.6                               & 50.2                                & —                                \\
$^{\ddagger}$Feature-based MLR$^{\clubsuit}$ \cite{liu2015enhancing} & 16.3                                            & 43.4                                            & 60.8                                             & —                                & 18.9                                            & 44.8                                            & 59.4                                             & —                                & 25.4                                            & 53.3                                            & 68.9                                             & —                                & 20.3                                            & 42.6                                            & \textbf{56.9}                                    & —                                & 31.8                               & 53.6                               & 64.1                                & —                                \\
AGE \cite{rao2020self}                    & 20.8                                            & 29.3                                            & 34.2                                             & 3.5                              & 37.1                                            & 56.2                                            & 67.0                                             & 9.8                              & 35.5                                            & 54.3                                            & 65.3                                             & 9.6                              & 14.6                                            & 33.0                                            & 42.7                                             & 3.0                              & 32.4                               & 51.2                               & 60.1                                & 3.9                              \\
SM-SGE \cite{rao2021sm}                   & 50.2                                            & 73.5                                            & 81.9                                             & 6.6                              & 26.6                                            & 49.0                                            & 59.4                                             & 9.3                              & 27.2                                            & 51.4                                            & 63.2                                             & 9.7                              & 10.6                                            & 26.3                                            & 35.9                                             & 3.0                              & 16.6                               & 36.8                               & 47.5                                & 3.5                              \\
SGELA \cite{rao2021self}                  & 71.8                                            & 87.5                                            & 91.4                                             & 9.8                              & 48.1                                            & 69.5                                            & 77.7                                             & 16.5                             & 51.2                                            & 73.8                                            & 81.5                                             & 7.1                              & 15.9                                            & 30.8                                            & 40.6                                             & 4.7                              & 36.4                               & 57.1                               & 64.6                                & 6.7                              \\
SimMC \cite{rao2022simmc}                 & 80.8                                            & 92.3                                            & 93.7                                             & 10.8                             & 69.1                                            & 86.6                                            & 91.3                                             & 16.5                             & 68.0                                            & 88.1                                            & \textbf{93.0}                                    & \textbf{15.7}                    & 25.6                                            & 43.8                                            & 54.0                                             & \textbf{5.4}                     & 42.0                               & 59.8                               & 68.9                                & 7.1                              \\
Hi-MPC$^{h}$ (Ours)                                        & \textbf{85.5}                                   & \textbf{94.9}                                   & \textbf{95.8}                                    & \textbf{11.2}                    & \textbf{71.2}                                   & \textbf{87.5}                                   & \textbf{92.1}                                    & \textbf{17.0}                    & \textbf{70.2}                                   & \textbf{88.5}                                   & 92.6                                             & 14.1                             & \textbf{27.2}                                   & \textbf{45.0}                                   & 54.9                                             & 4.9                              & \textbf{50.1}                      & \textbf{65.5}                      & \textbf{72.1}                       & \textbf{7.5}                     \\ \hline
\end{tabular}
}
}
\end{table*}

\begin{table*}[t]
\centering
\caption{Ablation study with different configurations: Direct prototype contrastive learning (DPC), meta-prototype contrastive learning (MPC), and hard skeleton mining mechanism (HSM). “Hi” denotes adopting hierarchical skeleton representations \mr{($i.e.$, combining skeleton-level, component-level and limb-level representations)} and exploiting the proposed MSMR for person re-ID. \mr{The configurations (ID = 1, 2, 3, 5) without using “Hi” adopt the original skeleton representation ($i.e.,$ joint-level representations).}  “+” indicates using the corresponding component. }
\label{ablation}
\scalebox{0.7}{
\setlength{\tabcolsep}{3.45mm}{
\begin{tabular}{clcccccccccccc}
\hline
\multirow{2}{*}{\textbf{ID}} & \multirow{2}{*}{\textbf{Configurations}} & \multicolumn{2}{c}{\textbf{BIWI-S}} & \multicolumn{2}{c}{\textbf{BIWI-W}} & \multicolumn{2}{c}{\textbf{IAS-A}} & \multicolumn{2}{c}{\textbf{IAS-B}} & \multicolumn{2}{c}{\textbf{KGBD}} & \multicolumn{2}{c}{\textbf{KS20}} \\
                             &                                          & \textbf{R$_{1}$}   & \textbf{mAP}   & \textbf{R$_{1}$}   & \textbf{mAP}   & \textbf{R$_{1}$}   & \textbf{mAP}  & \textbf{R$_{1}$}   & \textbf{mAP}  & \textbf{R$_{1}$}  & \textbf{mAP}  & \textbf{R$_{1}$}  & \textbf{mAP}  \\ \hline
1                            & Baseline                                 & 24.8               & 9.3            & 10.9               & 14.1           & 29.4               & 13.8          & 30.2               & 13.3          & 20.5              & 4.4           & 17.0              & 9.5           \\
2                            & + DPC                                     & 38.3               & 10.7           & 19.9               & 19.7           & 35.4               & 16.3          & 35.4               & 16.6          & 53.7              & 8.5           & 63.3              & 17.6          \\
3                            & + MPC                                    & 39.8               & 13.1           & 22.4               & 19.3           & 38.4               & 17.0          & 37.3               & 14.4          & 53.2              & 8.1           & 63.9              & 18.5          \\
4                            & + Hi + MPC                               & 40.4               & 12.8           & 24.2               & 21.1           & 42.2               & 19.4          & 38.2               & 18.9          & 55.3              & 8.7           & 66.4              & 18.6          \\
5                            & + MPC + HSM                              & 44.9               & 14.2           & 23.7               & 21.0           & 40.0               & 17.8          & 42.8               & 23.1          & 55.8              & 9.5           & 66.2              & 19.0          \\
6                            & + Hi + MPC + HSM                         & 47.5               & 17.4           & 27.3               & 22.6           & 45.6               & 23.2          & 48.2               & 25.3          & 56.9              & 10.2          & 69.6              & 22.0          \\ \hline
\end{tabular}
}
}
\end{table*}
\label{estimated_skeletons}
In this section, we verify the generality of our skeleton-based approach under the large-scale RGB scenarios (CASIA-B). We leverage pre-trained pose estimation models \cite{cao2019openpose,chen20173d} to extract 3D skeleton data from RGB videos of CASIA-B, and evaluate the performance of our approach with the estimated skeleton data. 
As shown in \sloppy{Table \ref{CASIA_B_result}}, we compare our model with state-of-the-art skeleton-based methods \cite{rao2020self,rao2021self,rao2021sm,rao2022simmc} and representative appearance-based methods \cite{weinberger2009distance,davis2007information,gray2008viewpoint,farenzena2010person,liu2015enhancing}.
Our approach achieves superior performance to most state-of-the-art skeleton-based methods (AGE \cite{rao2020self}, SM-SGE \cite{rao2021sm}, SGELA \cite{rao2021self}) with a significant margin of $11.3$ to $64.7\%$ \hc{for} Rank-1 accuracy and $0.2$ to $7.7\%$ \hc{for} mAP in different evaluation conditions of CASIA-B. Compared with the latest contrastive learning framework SimMC \cite{rao2022simmc}, our model performs better in most conditions, which justifies its effectiveness on learning more discriminative skeleton representations when applied to \mr{RGB-estimated} skeleton data. Our skeleton-based approach also \mr{achieves higher results than several} representative  classical appearance-based methods that employ visual metric learning with RGB-based appearances and textures (LMNN \cite{weinberger2009distance}, ITML \cite{davis2007information}, ELF \cite{gray2008viewpoint}, SDALF \cite{farenzena2010person}) or leverage both gait energy images and appearance features (MLR \cite{liu2015enhancing}) on different conditions. \mr{It should be noted that directly comparing skeleton-based method with appearance-based methods might be unfair as they use different learning manners ($e.g.,$ supervised or unsupervised learning) and fundamentally different data modalities ($e.g.,$ gait energy images). However, these classic representative methods can serve as a baseline performance reference under the same evaluation protocol \cite{liu2015enhancing,rao2022simmc}.}
The highly competitive performance of the proposed approach on RGB-estimated skeletons demonstrates its \hc{generality and value} for person re-ID under large-scale RGB-based scenarios.

\begin{table*}[t]
\centering
\caption{Performance of our approach on different datasets when solely exploiting limb-level (L), component-level (C), joint-level skeleton representations (J) or MSMR that combines all level skeleton representations for person re-ID. We compare our approach and the na\"ive Hi-MPC without using HSM. “$\checkmark$” indicates using the corresponding configurations.}
\label{graph_level}
\scalebox{0.7}{
\setlength{\tabcolsep}{3.3mm}{
\begin{tabular}{cccccrrrrrrrrrrrr}
\hline
\multirow{2}{*}{\textbf{ID}} & \multirow{2}{*}{\textbf{L}} & \multirow{2}{*}{\textbf{C}} & \multirow{2}{*}{\textbf{J}} & \multirow{2}{*}{\textbf{HSM}} & \multicolumn{2}{c}{\textbf{BIWI-S}}                               & \multicolumn{2}{c}{\textbf{BIWI-W}}                               & \multicolumn{2}{c}{\textbf{IAS-A}}                                & \multicolumn{2}{c}{\textbf{IAS-B}}                                 & \multicolumn{2}{c}{\textbf{KGBD}}                                 & \multicolumn{2}{c}{\textbf{KS20}}                                 \\
                             &                    &                             &                             &                               & \textbf{R$_{1}$} & \textbf{mAP} & \textbf{R$_{1}$} & \textbf{mAP} & \textbf{R$_{1}$} & \textbf{mAP} & \textbf{R$_{1}$} & \textbf{mAP}  & \textbf{R$_{1}$} & \textbf{mAP} & \textbf{R$_{1}$} & \textbf{mAP} \\ \hline
1                            & \checkmark                  &                             &                             &                               & 35.7                              & 12.1                          & 18.0                              & 16.8                          & 35.2                              & 16.1                          & 38.0                              & 18.3                           & 42.5                              & 5.5                          & 59.2                              & 15.6                          \\
2                            &                    & \checkmark                           &                             &                               & 38.9                              & 12.2                          & 21.8                              & 18.1                          & 36.9                              & 16.0                          & 38.0                              & 19.8                           & 50.0                              & 7.0                           & 59.8                              & 16.4                          \\
3                            &                    &                             & \checkmark                           &                               & 39.3                              & 12.3                          & 22.4                              & 19.3                          & 38.4                              & 17.0                          & 37.3                              & 14.4                           & 51.5                              & 7.7                           & 61.8                              & 16.5                          \\
4                            & \checkmark                  & \checkmark                           & \checkmark                           &                               & 40.4                              & 12.8                          & 24.2                              & 21.1                          & 42.2                              & 19.4                          & 38.2                              & 18.9                           & 55.3                              & 8.7                           & 65.4                              & 18.2                          \\
5                            & \checkmark                  &                             &                             & \checkmark                             & 36.9                              & 12.4                          & 17.5                              & 16.3                          & 35.5                              & 17.7                          & 40.1                              & 19.3                           & 44.3                              & 5.7                          & 64.1                              & 17.9                          \\
6                            &                    & \checkmark                           &                             & \checkmark                             & 41.8                              & 15.1                          & 22.3                              & 18.9                          & 41.7                              & 18.1                          & 44.6                              & 20.5                           & 52.3                              & 7.9                           & 66.0                              & 18.8                          \\
7                            &                    &                             & \checkmark                           & \checkmark                             & 44.9                              & 14.2                          & 23.7                              & 21.0                          & 40.0                              & 17.8                          & 41.7    & 20.1 & 55.8                              & 9.5                           & 65.4                              & 18.9                          \\
8                            & \checkmark                  & \checkmark                           & \checkmark                           & \checkmark                             & 47.5                              & 17.4                          & 27.3                              & 22.6                          & 45.6                              & 23.2                          & 48.2                              & 25.3                           & 56.9                              & 10.2                          & 69.6                              & 22.0                          \\ \hline
\end{tabular}
}
}
\end{table*}

\begin{figure*}[t]
    \centering
      \subfigure[Multi-Level]{\scalebox{0.3}{\includegraphics[]{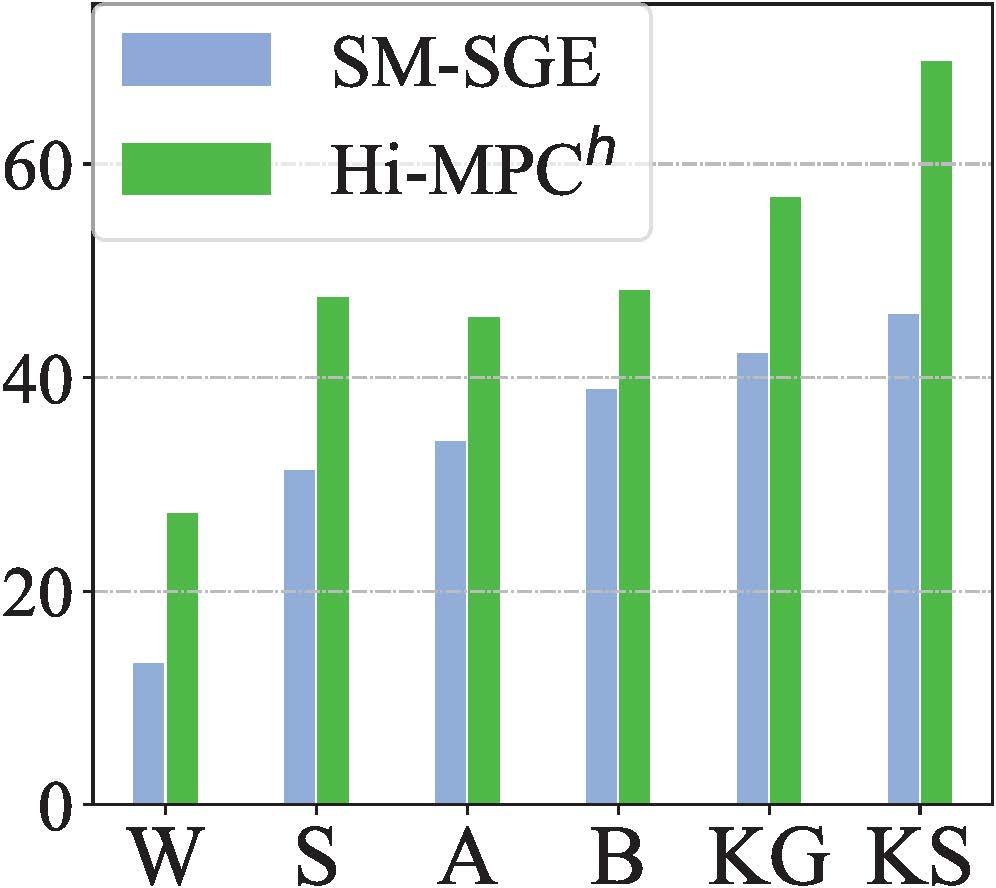}}} \
      \subfigure[Joint-Level]{\scalebox{0.3}{\includegraphics[]{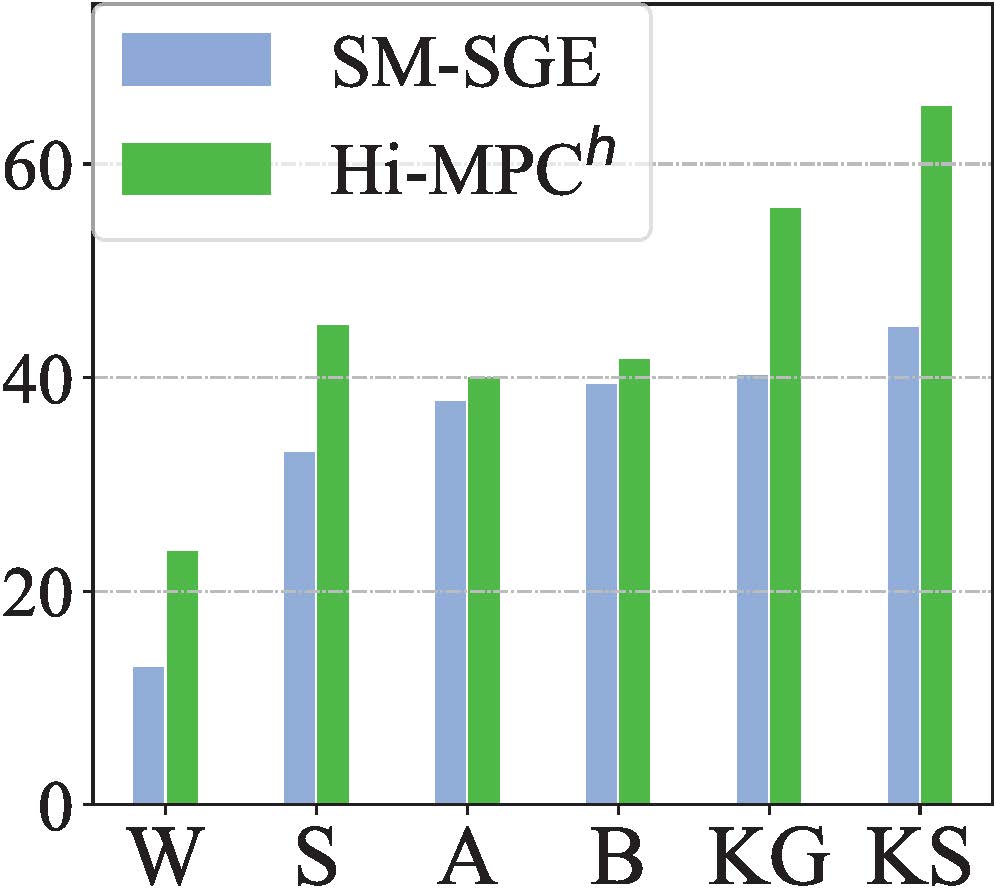}}} \
       \subfigure[Component-Level]{\scalebox{0.3}{\includegraphics[]{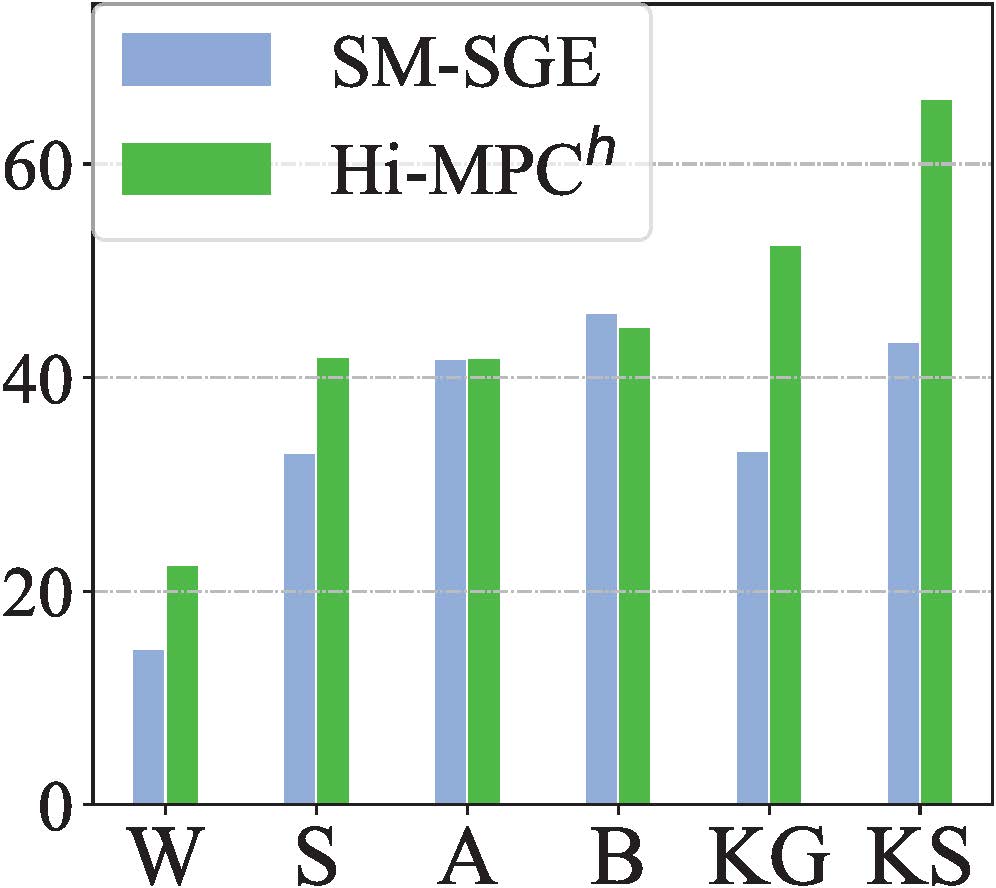}}} \
       \subfigure[Limb-Level]{\scalebox{0.3}{\includegraphics[]{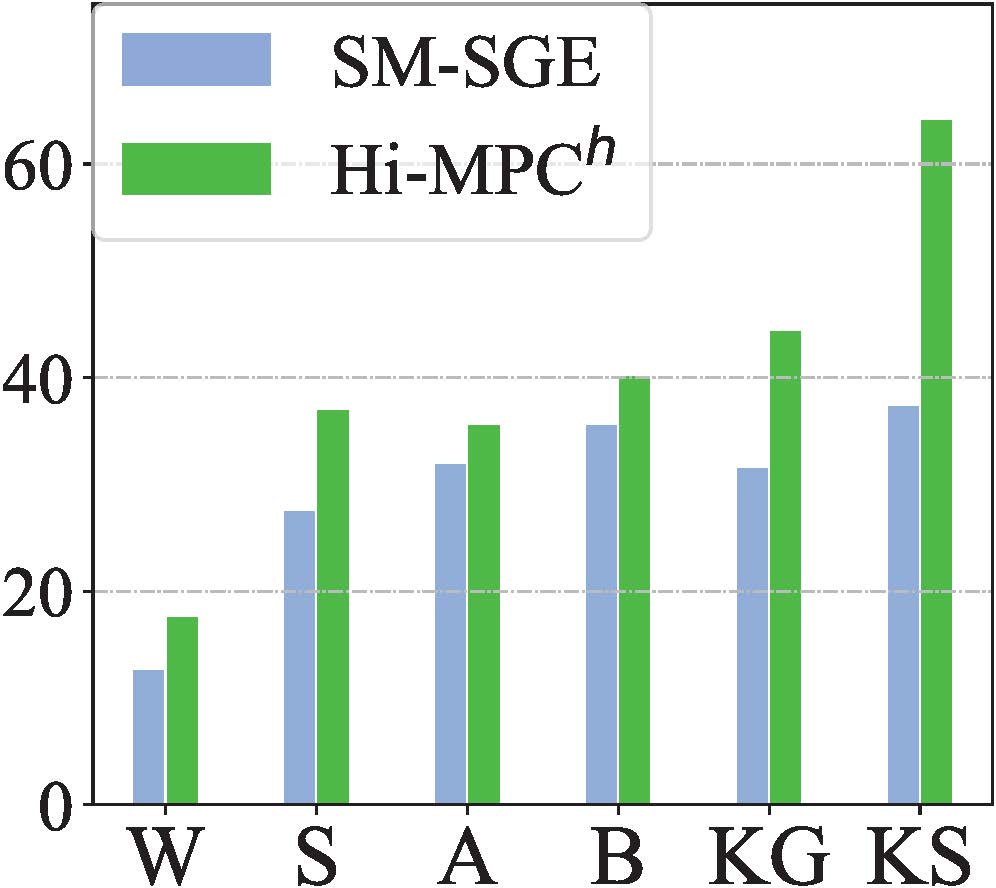}}} 
    \caption{Performance (Rank-1 accuracy) of our approach with different level representations (joint-level, component-level, limb-level) and their combination (multi-level) on BIWI-W (W), BIWI-S (S), IAS-A (A), IAS-B (B), KGBD (KG), and KS20 (KS) testing sets. The latest multi-level method SM-SGE \cite{rao2021sm} is compared under the same setting of skeleton levels.}
    \label{level_comp}
\end{figure*}

\subsection{Ablation Study}
We conduct ablation study to demonstrate the contribution of each component in our approach. The raw skeleton sequences, $i.e.,$ concatenation of 3D coordinates of body joints, are adopted as the baseline for comparison. As reported in \sloppy{Table \ref{ablation}}, employing direct skeleton prototype contrastive (DPC) learning (ID = 2) significantly improves the person re-ID performance of \hc{the} baseline especially on more challenging datasets such as KGBD (up to $33.2\%$ \hc{for} Rank-1 accuracy) and KS20 (up to $46.3\%$ \hc{for} Rank-1 accuracy). \hc{This} suggests that prototypes are indeed more representative skeleton features than raw 3D skeletons and DPC plays a crucial role in exploiting such discriminative features from different individuals. The proposed meta-prototype contrastive (MPC) learning (ID = 3) performs better than DPC (ID = 2) in almost all cases when applied to joint-level skeleton representations. \hc{Combining} hierarchical skeleton representations for multi-level clustering and meta-prototype contrastive learning (ID = 4) achieves \hc{better} performance than joint-level DPC (ID = 2) by up to $6.8\%$ \hc{for} Rank-1 accuracy and $3.1\%$ \hc{for} mAP on different datasets. Such results verify the effectiveness of the proposed Hi-MPC, as it can capture richer body and motion patterns via exploiting the most typical skeleton features from various levels. The effects of different level skeleton representations will be also discussed in Sec. \ref{dis_levels}. The proposed approach (Hi-MPC$^{h}$) employing the hard skeleton mining (HSM) mechanism (ID = 6) achieves consistent performance improvement of $1.6$-$10.0\%$ \hc{for} Rank-1 accuracy and $1.2$-$6.4\%$ \hc{for} mAP compared with na\"ive Hi-MPC (ID = 4) on all datasets. Adding HSM (ID = 5) also \hc{improves} the performance of MPC (ID = 3), which suggests the general validity of HSM 
for both single-level and hierarchical skeleton contrastive learning. These results further \hc{suggest} that HSM can facilitate mining key skeletons to learn highly informative and valuable skeletal patterns during skeleton meta-prototype contrastive learning for person re-ID. Further visualization and analysis of the HSM mechanism are provided in Sec. \ref{analysis_HSM}.

\begin{table*}[t]
\centering
\caption{Performance of our approach with different meta-transformation heads ($M$ = 1, 4, 8, 16).}
\label{dis_M}
\scalebox{0.7}{
\setlength{\tabcolsep}{4.99mm}{
\begin{tabular}{crrrrrrrrrrrr}
\hline
\multirow{2}{*}{\textbf{$M$}} & \multicolumn{2}{c}{\textbf{BIWI-S}} & \multicolumn{2}{c}{\textbf{BIWI-W}} & \multicolumn{2}{c}{\textbf{IAS-A}} & \multicolumn{2}{c}{\textbf{IAS-B}} & \multicolumn{2}{c}{\textbf{KS20}} & \multicolumn{2}{c}{\textbf{KGBD}} \\
                              & \textbf{R$_{1}$}   & \textbf{mAP}   & \textbf{R$_{1}$}   & \textbf{mAP}   & \textbf{R$_{1}$}   & \textbf{mAP}  & \textbf{R$_{1}$}   & \textbf{mAP}  & \textbf{R$_{1}$}  & \textbf{mAP}  & \textbf{R$_{1}$}  & \textbf{mAP}  \\ \hline
1                             & 45.7               & 15.3           & 25.7               & 22.5           & 43.8               & 21.1          & 46.7               & 25.0          & 67.2              & 20.6          & 55.5              & 9.2           \\
4                             & 47.1               & 17.6           & 26.6               & 22.9           & 45.0               & 22.5          & 48.1               & 26.8          & 69.2              & 21.3          & 56.5              & 10.0          \\
8                             & 47.5               & 17.4           & 27.3               & 22.6           & 45.6               & 23.2          & 48.2               & 25.3          & 69.6              & 22.0          & 56.9              & 10.2          \\
16                            & 48.5               & 16.9           & 27.7               & 23.9           & 46.0               & 24.0          & 48.2               & 25.4          & 70.7              & 21.4          & 57.0              & 9.7           \\ \hline
\end{tabular}
}
}
\end{table*}

\begin{table*}[t]
\centering
\caption{Performance of our approach with different embedding sizes ($h$ = 64, 128, 256, 512).}
\label{dis_H}
\scalebox{0.7}{
\setlength{\tabcolsep}{4.93mm}{
\begin{tabular}{crrrrrrrrrrrr}
\hline
\multirow{2}{*}{\textbf{$h$}} & \multicolumn{2}{c}{\textbf{BIWI-S}} & \multicolumn{2}{c}{\textbf{BIWI-W}} & \multicolumn{2}{c}{\textbf{IAS-A}} & \multicolumn{2}{c}{\textbf{IAS-B}} & \multicolumn{2}{c}{\textbf{KS20}} & \multicolumn{2}{c}{\textbf{KGBD}} \\
                              & \textbf{R$_{1}$}   & \textbf{mAP}   & \textbf{R$_{1}$}   & \textbf{mAP}   & \textbf{R$_{1}$}   & \textbf{mAP}  & \textbf{R$_{1}$}   & \textbf{mAP}  & \textbf{R$_{1}$}  & \textbf{mAP}  & \textbf{R$_{1}$}  & \textbf{mAP}  \\ \hline
64                            & 45.1               & 16.3           & 25.1               & 21.6           & 44.0               & 21.7          & 46.5               & 24.5          & 69.2              & 21.4          & 54.7              & 8.7           \\
128                           & 47.0               & 17.5           & 27.5               & 22.9           & 45.0               & 22.9          & 49.3               & 26.9          & 69.7              & 20.8          & 55.1              & 9.5           \\
256                           & 47.5               & 17.4           & 27.3               & 22.6           & 45.6               & 23.2          & 48.2               & 25.3          & 69.6              & 22.0          & 56.9              & 10.2          \\
512                           & 47.0               & 15.7           & 26.2               & 22.3           & 42.5               & 21.8          & 49.5               & 27.3          & 70.7              & 22.0          & 55.8              & 9.2           \\ \hline
\end{tabular}
}
}
\end{table*}

\subsection{Evaluation on Different Level Skeleton Representations}
\label{dis_levels}
We evaluate the performance of limb-level, component-level, joint-level skeleton representations and their combination in the proposed approach. As shown in \sloppy{Table \ref{graph_level}}, different level skeleton representations (ID = 5-7) learned by our final approach can \textit{individually} achieve highly competitive performance on different datasets, which suggests the great potential of higher level skeleton representations such as key limbs to be directly applied for person re-ID. Notably, the model exploiting joint-level (ID = 3, 7) and component-level skeleton representations (ID = 2, 6) achieves higher performance than using limb-level (ID = 1, 5) in most cases. This demonstrates the greater contribution of low-level skeleton representations, as they usually contain more specific positional and structural information of body parts than high-level ones to benefit learning more discriminative features for person re-ID.
Furthermore, combining hierarchical skeleton representations of all levels (ID = 4, 8) attains the best person re-ID performance on different datasets when compared to single-level representations (ID = 1-3, 5-7), regardless of using HSM. Such results further verify the necessity of the proposed hierarchical skeleton representations, as the multi-level skeletal modeling can encourage mining more unique body and motion features for person re-ID, which is consistent with the analysis in \cite{rao2021multi,rao2021sm}.

We also compare \hc{the} performance of different-level skeleton representations learned by our approach with the latest multi-level graph framework SM-SGE \cite{rao2021sm}. As presented in Fig. \ref{level_comp}, our approach not only significantly outperforms SM-SGE using multi-level representations, but also gains superior performance when applying the learned higher level representations for person re-ID on five of six testing sets. \hc{This} further demonstrates the effectiveness of our approach on learning more useful skeleton features at various levels. \hc{It is interesting to observe that} the component-level representations with a simpler body structure can perform comparably or even better than joint-level representations on half of datasets. \hc{This implies that the original skeletons} of those datasets might contain redundant positional or structural information, which can be compressed and characterized with more concise and abstract skeleton representations to better achieve person re-ID.

\subsection{Discussions}
\subsubsection{Effects of Meta-Transformation Heads}
\sloppy{Table \ref{dis_M}} presents the effects of different numbers of meta-transformation heads. Employing more learnable meta-transformation heads is shown to improve the performance of our approach on different datasets, which demonstrates the effectiveness of exploiting different contrastive subspaces for better skeleton representation learning. \hc{It is worth noting that the model performance tends to be stable when applying many more heads}, as it could introduce more random perturbation into contrastive learning and help obtain more robust prototype estimation (see Sec. \ref{Hi-HPC_sec}). More analysis is provided in \rhc{the} Appendices.

\subsubsection{Effects of Embedding Sizes}
As shown in \sloppy{Table \ref{dis_H}}, our approach achieves higher performance with relatively larger embedding sizes ($h\geq128$) on all datasets. The results suggest that a small embedding size ($h=64$) could be insufficient to learn effective skeleton representations for person re-ID, while using too large sizes of embedding ($e.g.,$ $h=512$) might cause the model to learn more redundant feature information and degrade the overall performance.

\subsubsection{Different Settings of DBSCAN}
\begin{table*}[t]
\centering
\caption{Performance of our approach when setting different minimum sample amount ($a_{min}$ = 1, 2, 3, 4) for DBSCAN.}
\label{dis_a}
\scalebox{0.7}{
\setlength{\tabcolsep}{4.85mm}{
\begin{tabular}{crrrrrrrrrrrr}
\hline
\multirow{2}{*}{\textbf{$a_{min}$}} & \multicolumn{2}{c}{\textbf{BIWI-S}} & \multicolumn{2}{c}{\textbf{BIWI-W}} & \multicolumn{2}{c}{\textbf{IAS-A}} & \multicolumn{2}{c}{\textbf{IAS-B}} & \multicolumn{2}{c}{\textbf{KS20}} & \multicolumn{2}{c}{\textbf{KGBD}} \\
                                    & \textbf{R$_{1}$}   & \textbf{mAP}   & \textbf{R$_{1}$}   & \textbf{mAP}   & \textbf{R$_{1}$}   & \textbf{mAP}  & \textbf{R$_{1}$}   & \textbf{mAP}  & \textbf{R$_{1}$}  & \textbf{mAP}  & \textbf{R$_{1}$}  & \textbf{mAP}  \\ \hline
1                                   & 46.6               & 17.6           & 26.6               & 21.0           & 44.7               & 22.1          & 48.5               & 26.1          & 68.4              & 21.1          & 54.7              & 9.6           \\
2                                   & 47.5               & 17.4           & 27.3               & 22.6           & 45.6               & 23.2          & 48.2               & 25.3          & 69.6              & 22.0          & 56.3              & 10.0          \\
3                                   & 48.8               & 18.1           & 27.7               & 23.2           & 44.8               & 22.7          & 49.0               & 25.7          & 70.3              & 21.7          & 56.9              & 10.2          \\
4                                   & 46.9               & 17.8           & 27.6               & 23.3           & 45.4               & 23.2          & 49.1               & 26.1          & 69.0              & 20.8          & 56.3              & 10.2          \\ \hline
\end{tabular}
}
}
\end{table*}
We evaluate the effects of \hc{the} two main parameters in the DBSCAN algorithm, $i.e.,$ minimum sample amount $a_{min}$ within the maximum distance $\epsilon$, which are empirically selected to encourage more balanced and stable clustering. As presented in \sloppy{Table \ref{dis_a}}, $a_{min}$ seems to \hc{have} no significant effect on the performance of our approach, as setting different values of $a_{min}$ achieve similar accuracy on most datasets. Nevertheless, it is worth mentioning that employing \rhc{too small a value} for $a_{min}$ tends to generate much larger clusters and might lead to \hc{a degeneration of clustering ($e.g.,$ single super cluster)} and unstable model training in practice. 

\hc{Large} values of $\epsilon$ ($e.g.,$ $\epsilon=1.0$) \hc{greatly} reduce the performance of our approach, as shown in \sloppy{Table \ref{dis_eps}}. Considering that larger $\epsilon$ leads to higher connectedness of instances, $i.e.,$ larger cluster and smaller number of prototypes, the results demonstrate that setting \hc{a relatively lower $\epsilon$ value} with more diverse skeleton prototypes could facilitate learning richer discriminative features for person re-ID. However, \rhc{too small a value for} $\epsilon$ could cause excessive over-clustering, which \hc{leads} to the degradation of model performance on some datasets.


\begin{table*}[t]
\centering
\caption{Performance of our approach when setting different maximum distances ($\epsilon$ = 0.4, 0.6, 0.8, 1.0) for DBSCAN.}
\label{dis_eps}
\scalebox{0.7}{
\setlength{\tabcolsep}{4.95mm}{
\begin{tabular}{crrrrrrrrrrrr}
\hline
\multirow{2}{*}{\textbf{$\epsilon$}} & \multicolumn{2}{c}{\textbf{BIWI-S}} & \multicolumn{2}{c}{\textbf{BIWI-W}} & \multicolumn{2}{c}{\textbf{IAS-A}} & \multicolumn{2}{c}{\textbf{IAS-B}} & \multicolumn{2}{c}{\textbf{KS20}} & \multicolumn{2}{c}{\textbf{KGBD}} \\
                                     & \textbf{R$_{1}$}   & \textbf{mAP}   & \textbf{R$_{1}$}   & \textbf{mAP}   & \textbf{R$_{1}$}   & \textbf{mAP}  & \textbf{R$_{1}$}   & \textbf{mAP}  & \textbf{R$_{1}$}  & \textbf{mAP}  & \textbf{R$_{1}$}  & \textbf{mAP}  \\ \hline
0.4                                  & 46.8               & 16.6           & 19.9               & 18.7           & 43.2               & 18.2          & 42.0               & 20.0          & 69.5              & 20.7          & 55.6              & 7.8           \\
0.6                                  & 47.5               & 17.4           & 22.4               & 19.2           & 44.7               & 20.7          & 46.7               & 25.8          & 69.4              & 21.8          & 56.9              & 10.2          \\
0.8                                  & 50.3               & 18.2           & 27.3               & 22.6           & 45.6               & 23.2          & 48.2               & 25.3          & 69.6              & 22.0          & 47.9              & 4.8           \\
1.0                                  & 27.9               & 9.7            & 11.6               & 14.2           & 33.8               & 13.6          & 37.2               & 14.3          & 50.0              & 11.0          & 46.9              & 4.3           \\ \hline
\end{tabular}
}
}
\end{table*}



\subsubsection{Effects of Sequence Lengths}
\begin{figure*}[t]
    \centering
      \subfigure[$F=4$]{\scalebox{0.27}{\includegraphics[]{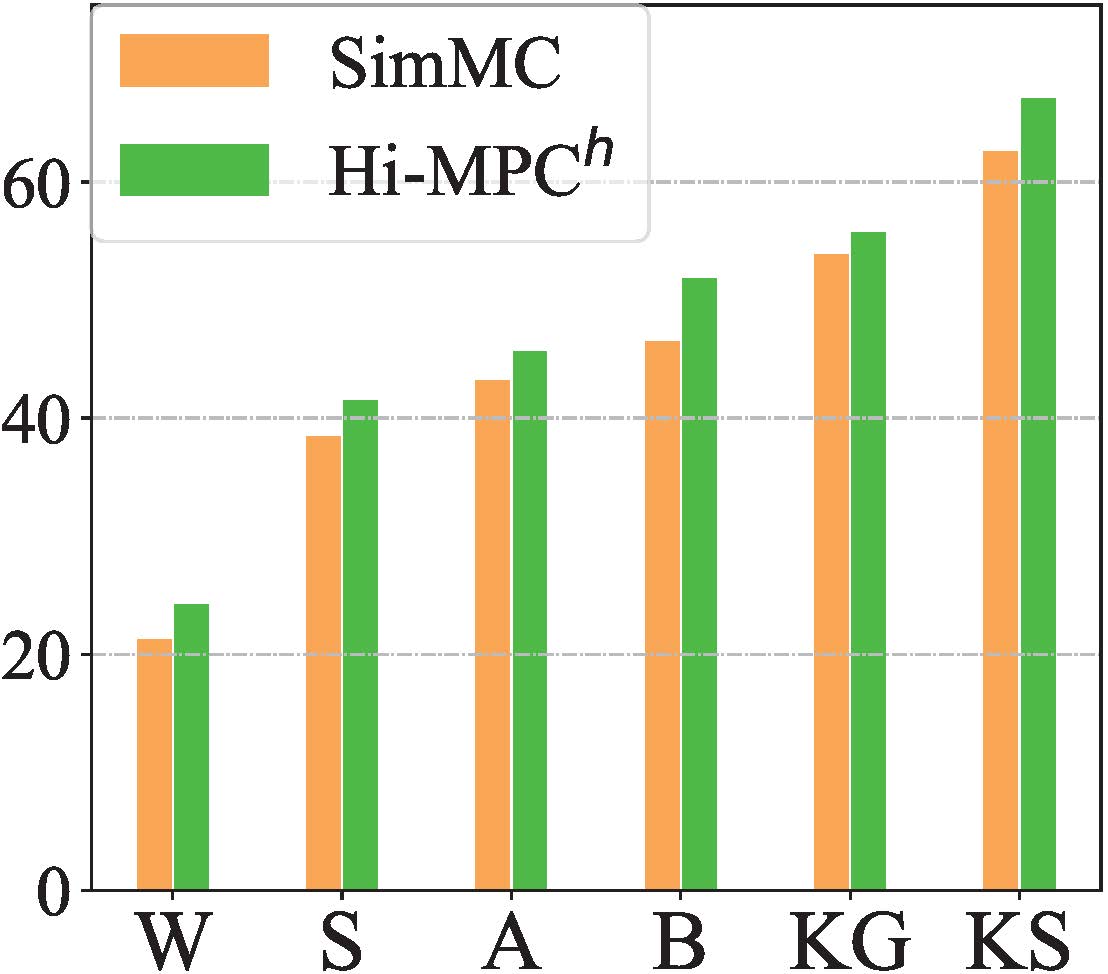}}} \
      \subfigure[$F=6$]{\scalebox{0.27}{\includegraphics[]{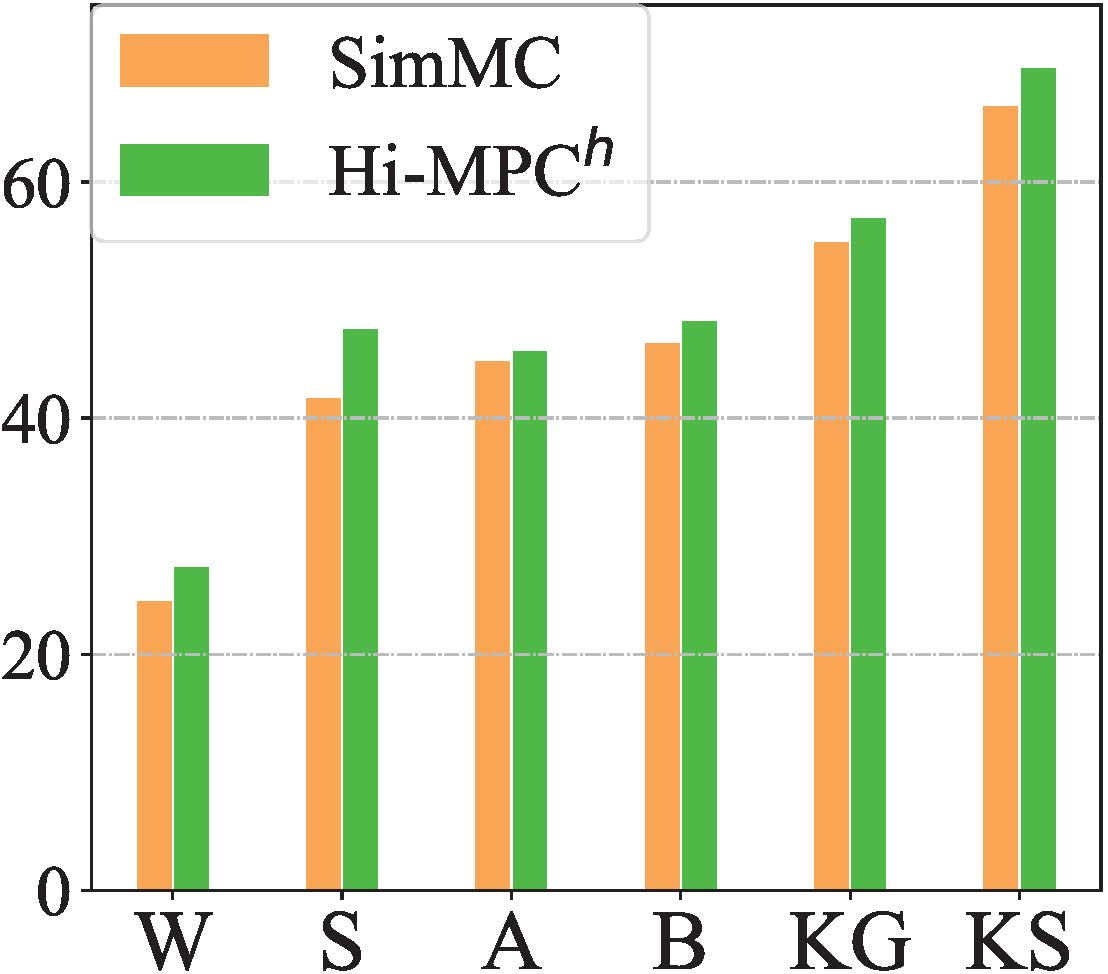}}} \
       \subfigure[$F=8$]{\scalebox{0.27}{\includegraphics[]{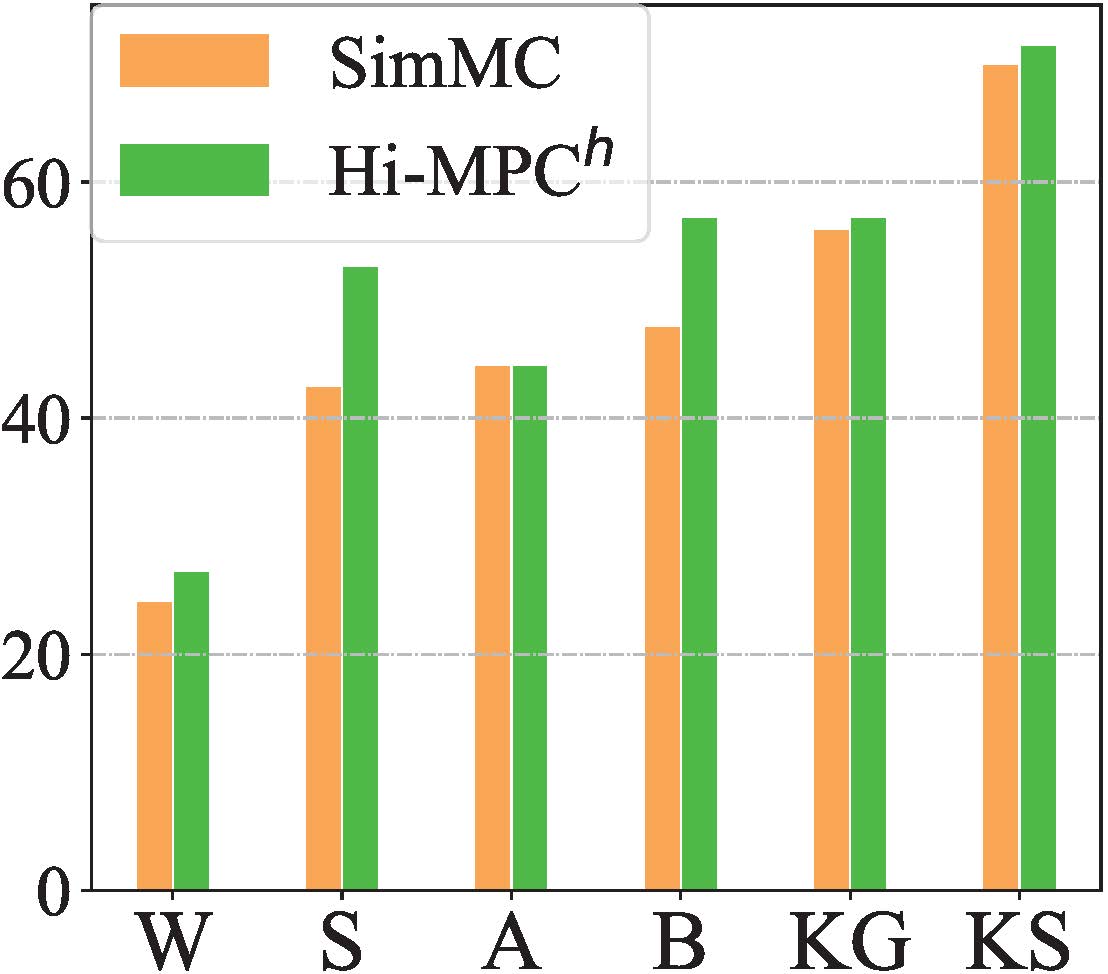}}} \
       \subfigure[$F=10$]{\scalebox{0.27}{\includegraphics[]{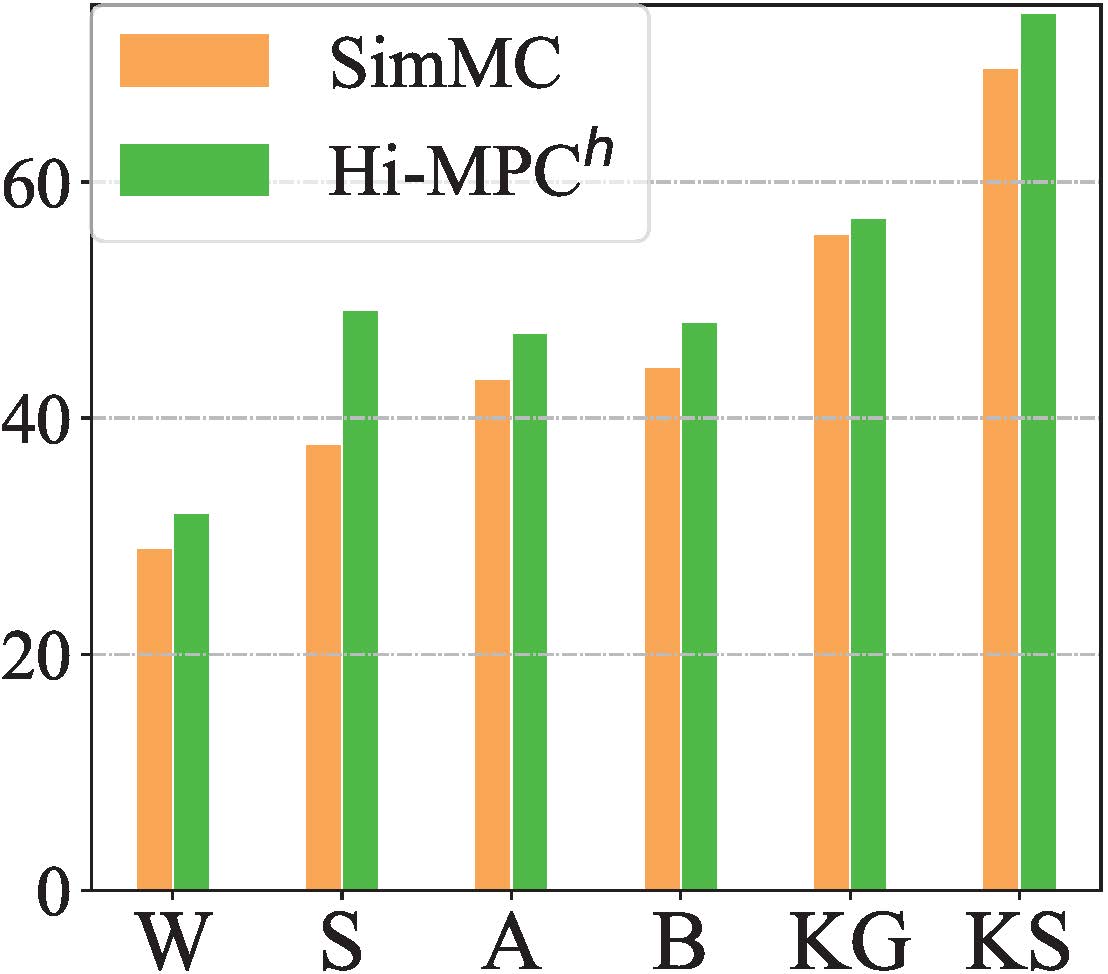}}} 
    \caption{Multi-shot person re-ID performance (Rank-1 accuracy) of our approach with different settings of sequence lengths ($F=4,6,8,10$) on BIWI-W (W), BIWI-S (S), IAS-A (A), IAS-B (B), KGBD (KG), and KS20 (KS) testing sets. The latest state-of-the-art method SimMC \cite{rao2022simmc} is compared as the performance baseline.}
    \label{F_comp}
\end{figure*}
We evaluate multi-shot performance of our approach with different settings of sequence lengths $F$ ($i.e.,$ $F$-shot person re-ID), and compare it with the latest SimMC \cite{rao2022simmc} on different datasets. As shown in Fig. \ref{F_comp}, the proposed Hi-MPC$^{h}$ consistently outperforms SimMC on all cases of datasets and sequence lengths. In contrast to SimMC that fails to keep high accuracy when $F$ \hc{varies slightly, our approach is more stable} with better results on different datasets. \hc{Since} skeleton sequences contain more pattern features \hc{as $F$ increases}, our approach is capable of learning more effective skeleton representations to achieve larger performance improvement in most cases. Nevertheless, it is \hc{interesting to note} that using shorter sequences sometimes performs better than longer sequences on small datasets such as IAS-B, implying that a larger size of available training sequences under smaller $F$ settings could help learn better representations on those datasets.

\subsection{Analysis of Hard Skeleton Mining}
\label{analysis_HSM}
\begin{figure}[t]
    \centering
         \scalebox{0.17}{\includegraphics{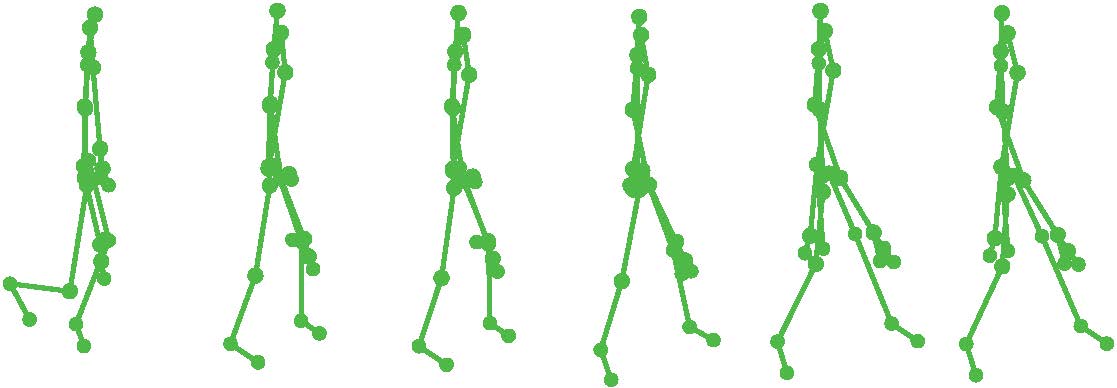}} 
           \ \ \scalebox{0.17}{\includegraphics{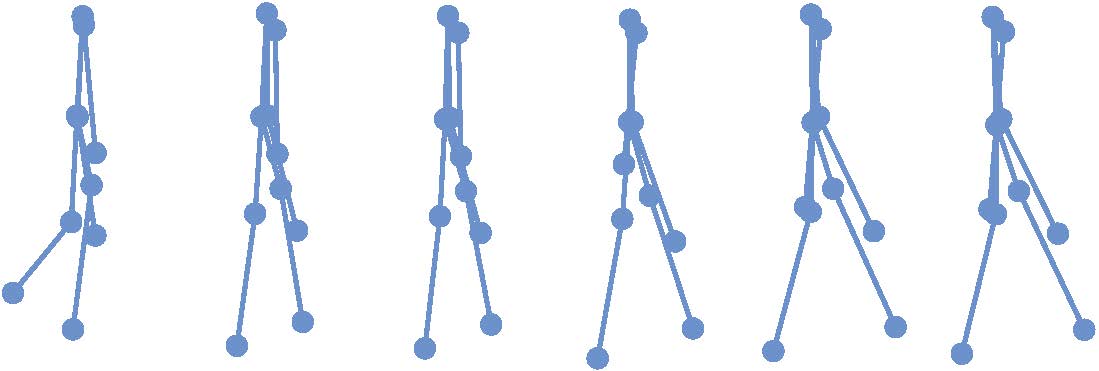}}   
          \ \
         \scalebox{0.17}{\includegraphics{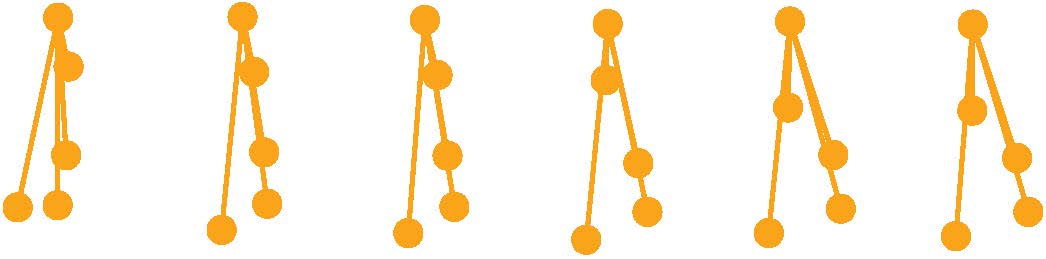}} 
         \\
          \scalebox{0.5}{\includegraphics{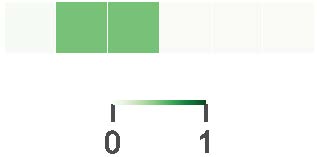}} 
           \quad \   \scalebox{0.5}{\includegraphics{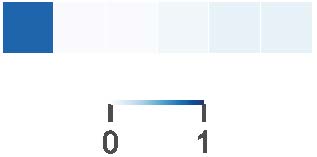}} 
          \quad  \
         \scalebox{0.5}{\includegraphics{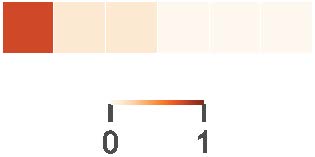}} 
         \\
           \scalebox{0.12}{\includegraphics{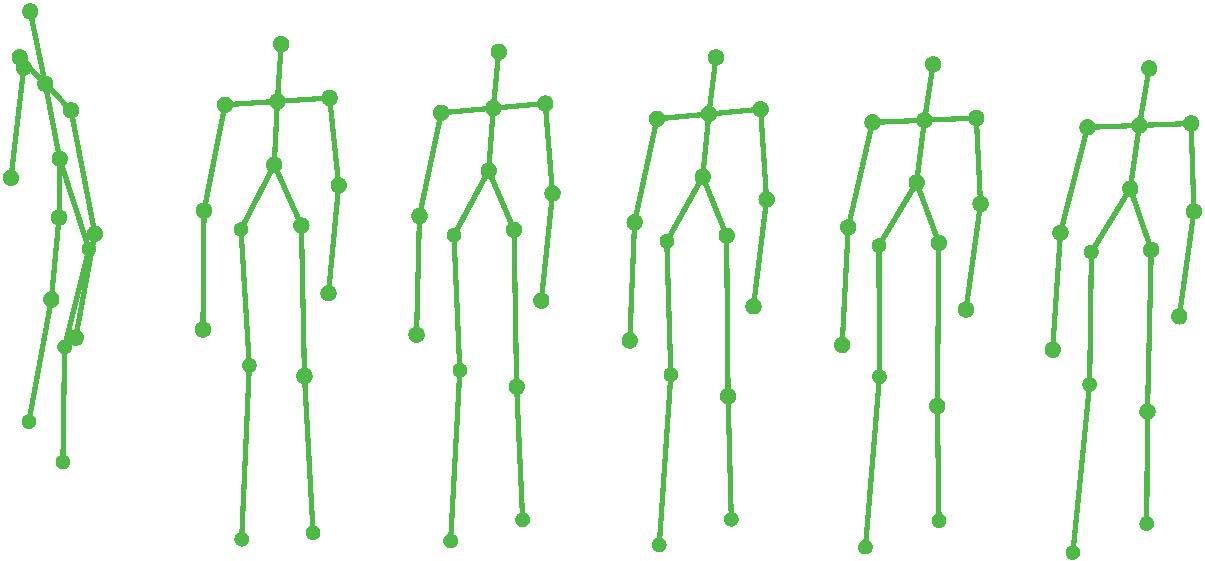}} 
           \ \quad \ \scalebox{0.12}{\includegraphics{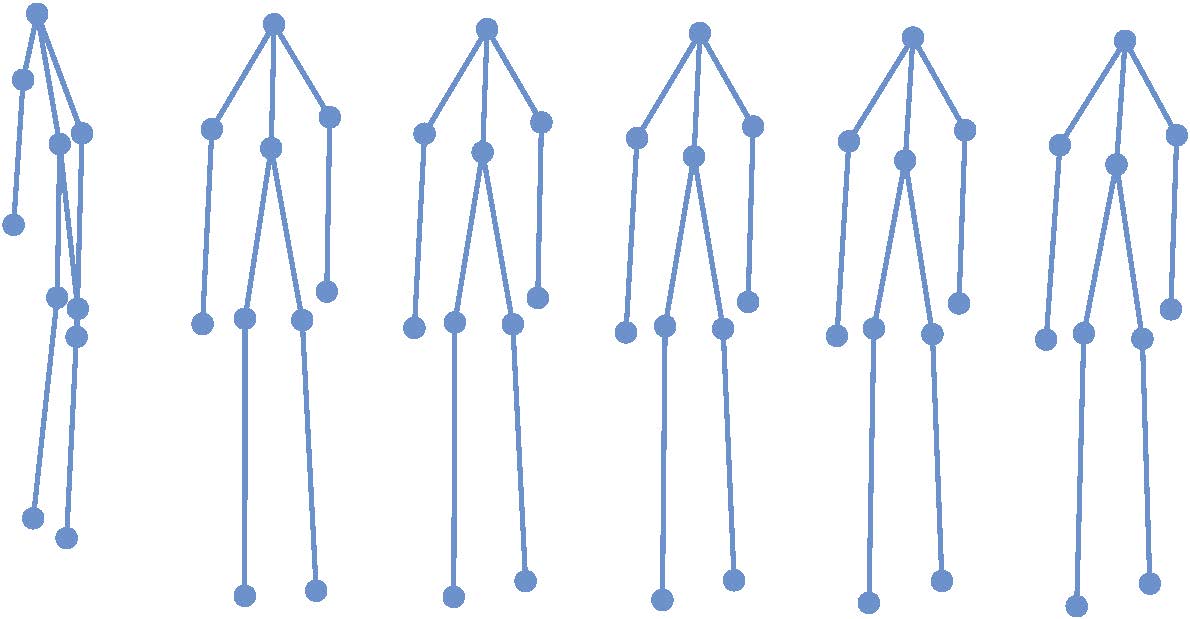}}   
          \ \quad \ 
         \scalebox{0.12}{\includegraphics{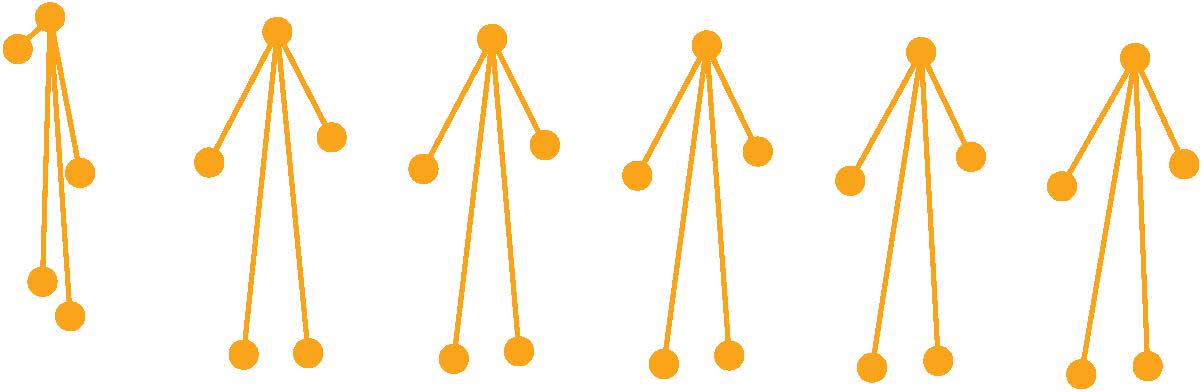}} 
         \\
          \scalebox{0.5}{\includegraphics{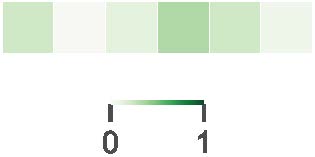}} 
           \quad \ \scalebox{0.5}{\includegraphics{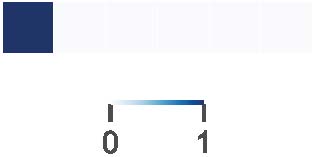}}
          \quad \  
         \scalebox{0.5}{\includegraphics{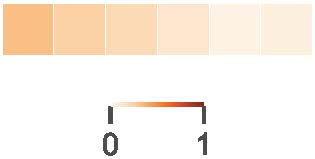}}
          \\
           \scalebox{0.14}{\includegraphics{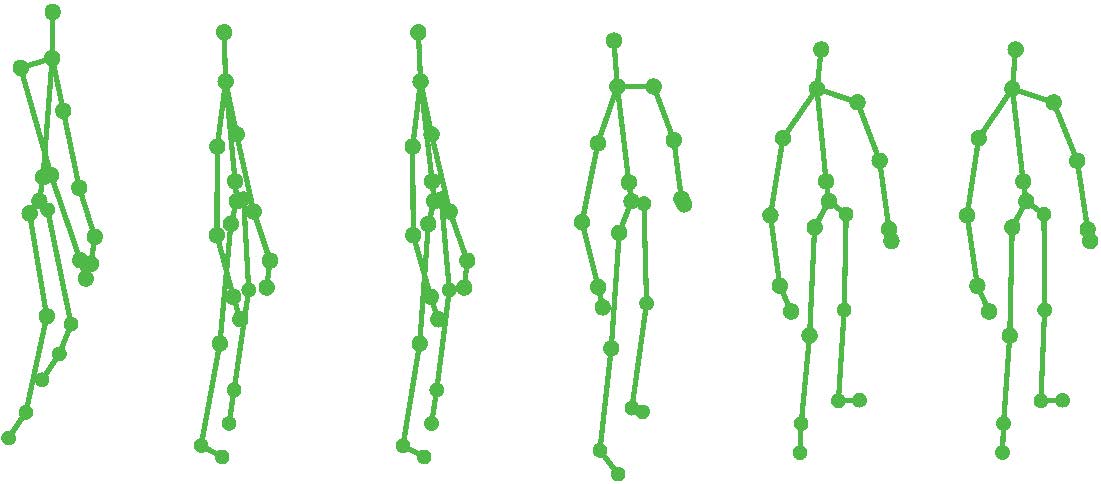}} 
            \quad  \scalebox{0.14}{\includegraphics{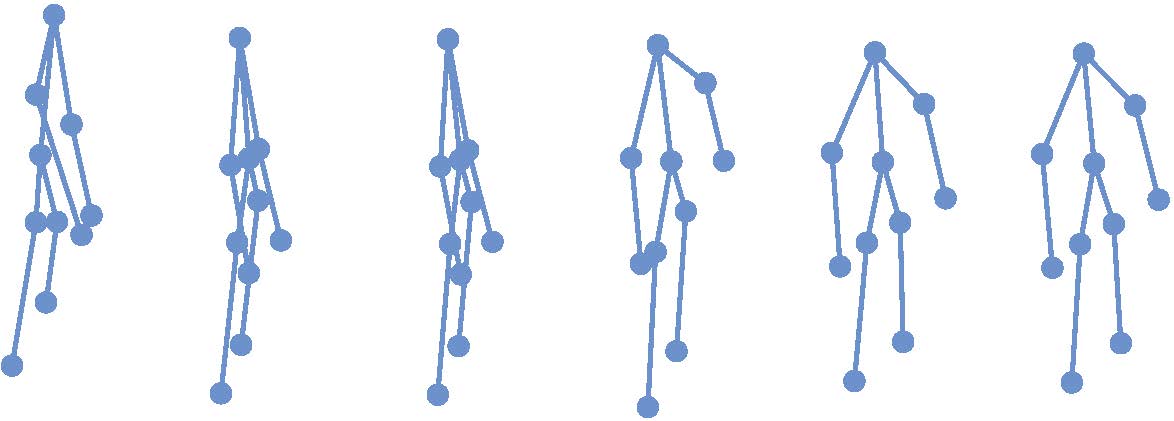}}   
          \quad 
         \scalebox{0.14}{\includegraphics{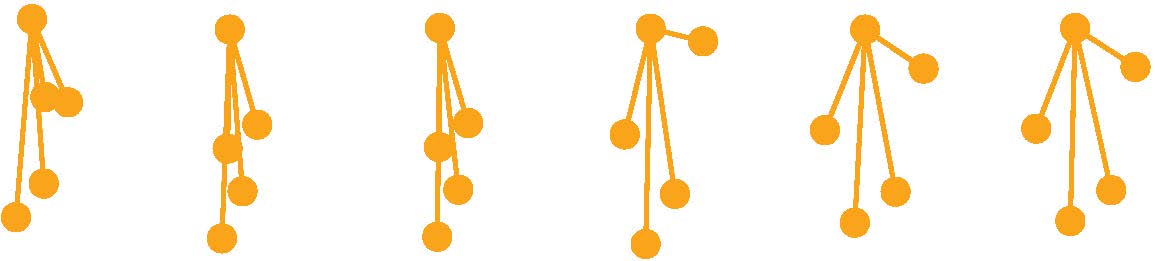}} 
         \\
          \scalebox{0.5}{\includegraphics{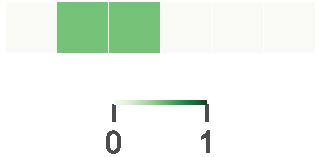}} 
           \quad \  \scalebox{0.5}{\includegraphics{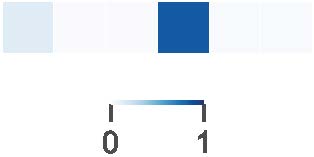}}
          \quad \
         \scalebox{0.5}{\includegraphics{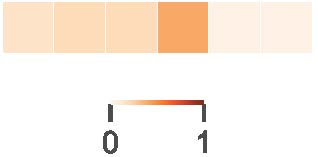}} 
    \caption{\hc{Visualization of joint-level (green), component-level (blue), limb-level representations (orange) of consecutive skeletons and their informative importance in different datasets. Each row shows three level representations of the same sequence. Darker colors of $i^{th}$ position in heat maps indicate higher importance of $i^{th}$ skeleton representation. }}
    \label{vis_HSM}
\end{figure}

To verify the effectiveness of the proposed HSM mechanism on mining more informative skeletons, we visualize joint-level, component-level, and limb-level skeleton representations with their inferred importance on different datasets. As shown in Fig. \ref{vis_HSM}, the higher importance is often assigned to either evidently different skeletons or dramatically changing poses in the sequence, which could introduce more pattern information compared with other similar skeletons. \hc{This is consistent with our intuition that skeletons containing} diverse patterns ($i.e.,$ higher intra-class variation) are harder to be recognized as the same person, thus deserving more attention to learn during training. It is also observed that there exists a good alignment of informative importance between component-level and limb-level skeleton representations, while our model focuses on more key skeletons with continuous patterns at joint-level, which suggests that the proposed approach may hierarchically capture different skeleton semantics ($e.g.,$ pattern consistency) to mine more effective features from various levels. 

\hc{The proposed HSM mechanism allows us to intuitively visualize the importance/value of each skeleton in learning \textit{hard} patterns and useful features. As shown in Fig. \ref{vis_HSM}, the component-level skeleton representations with a simpler body structure \rhc{are} highly similar with the joint-level skeleton representations to effectively characterize body poses, while the limb-level ones can provide more global motion dynamics of skeletons. This further suggests the potential of more concise and abstract skeleton representations on learning unique patterns for person re-ID. Moreover, as hard skeletons typically contain easily-confused or uncommon patterns, we can potentially exploit hard skeleton mining to detect special skeletons/poses ($e.g.,$ abnormal or pathological gaits) of a certain person for more advanced tasks such as medical gait analysis. It \rhc{could} also be used to discover noisy skeletons with incomplete or extremely unnatural poses for efficient skeleton filtration and selection.} 

\begin{figure*}[t]
    \centering
      \subfigure[SM-SGE \cite{rao2021sm}]{\scalebox{0.205}{\label{SMSGE_1_10_BIWI}\includegraphics[]{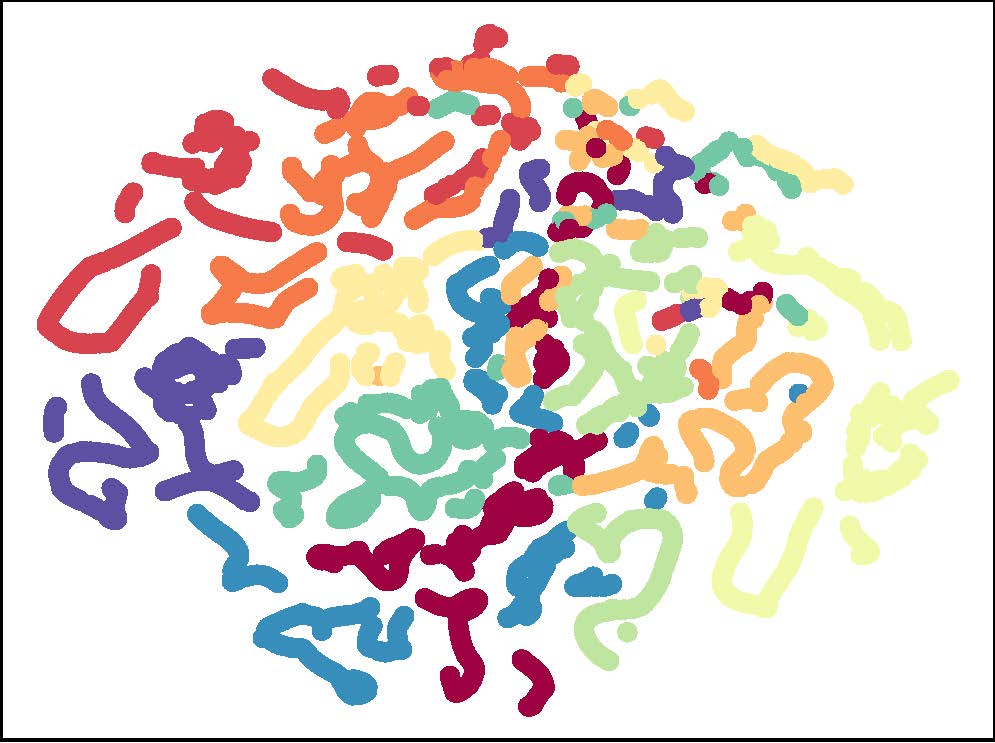}}} 
      \subfigure[SimMC \cite{rao2022simmc}]{\scalebox{0.205}{\label{SimMC_BIWI}\includegraphics[]{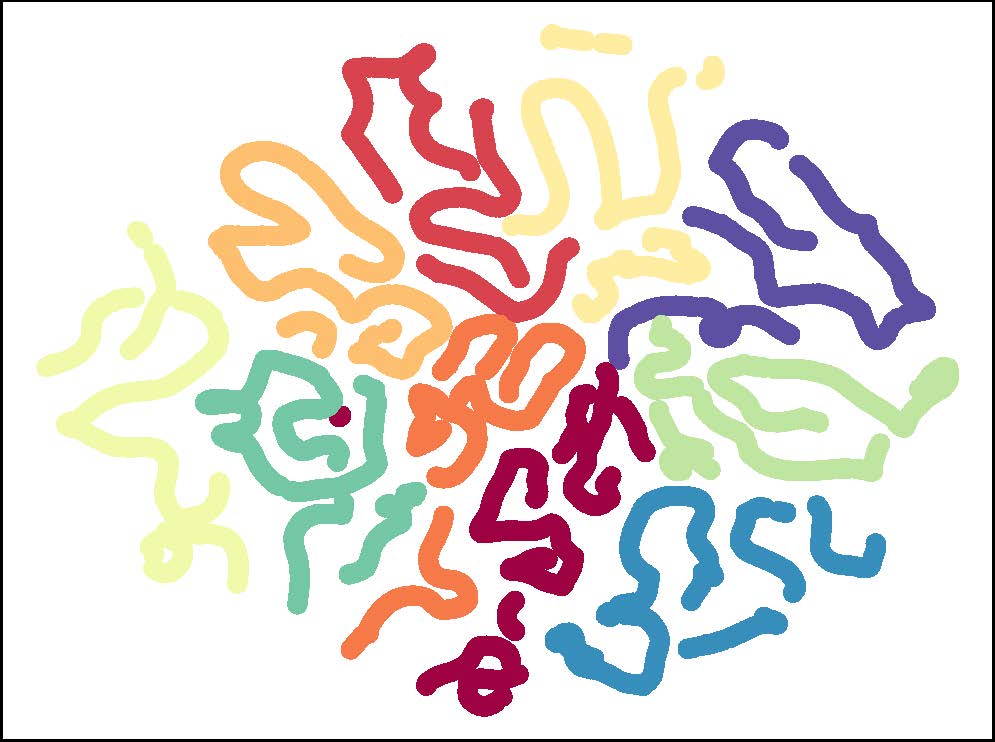}}} 
       \subfigure[Hi-MPC$^{h}$ (M)]{\scalebox{0.205}{\label{HiMPC_1_10_BIWI}\includegraphics[]{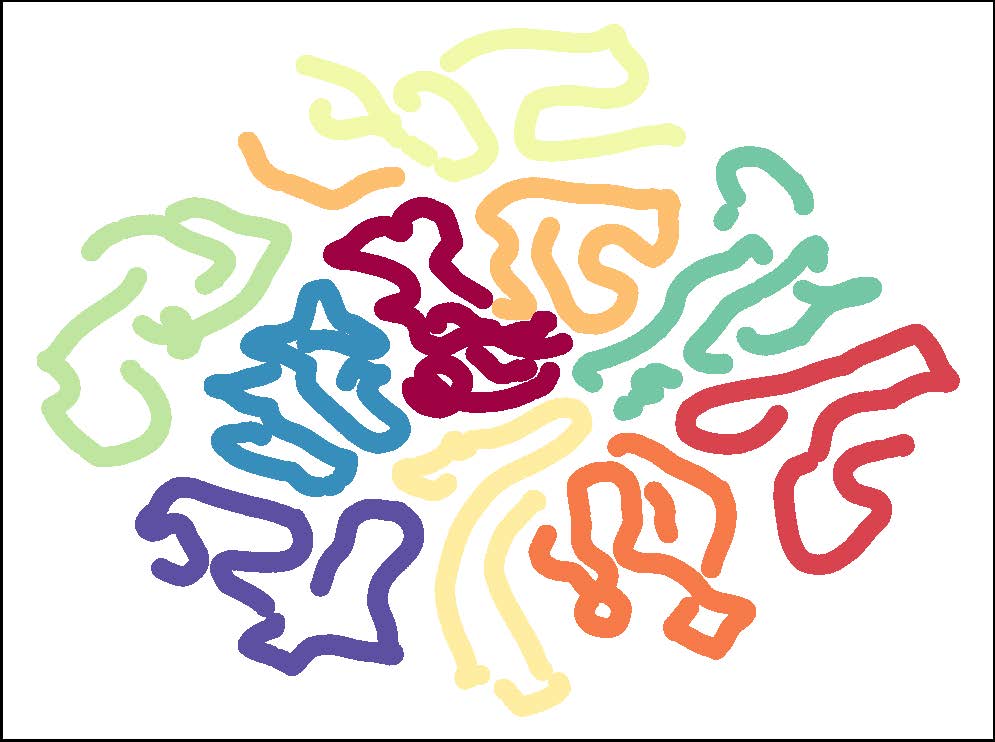}}} 
       \subfigure[Hi-MPC$^{h}$ (J)]{\scalebox{0.205}{\label{HiMPC_1_10_BIWI_J}\includegraphics[]{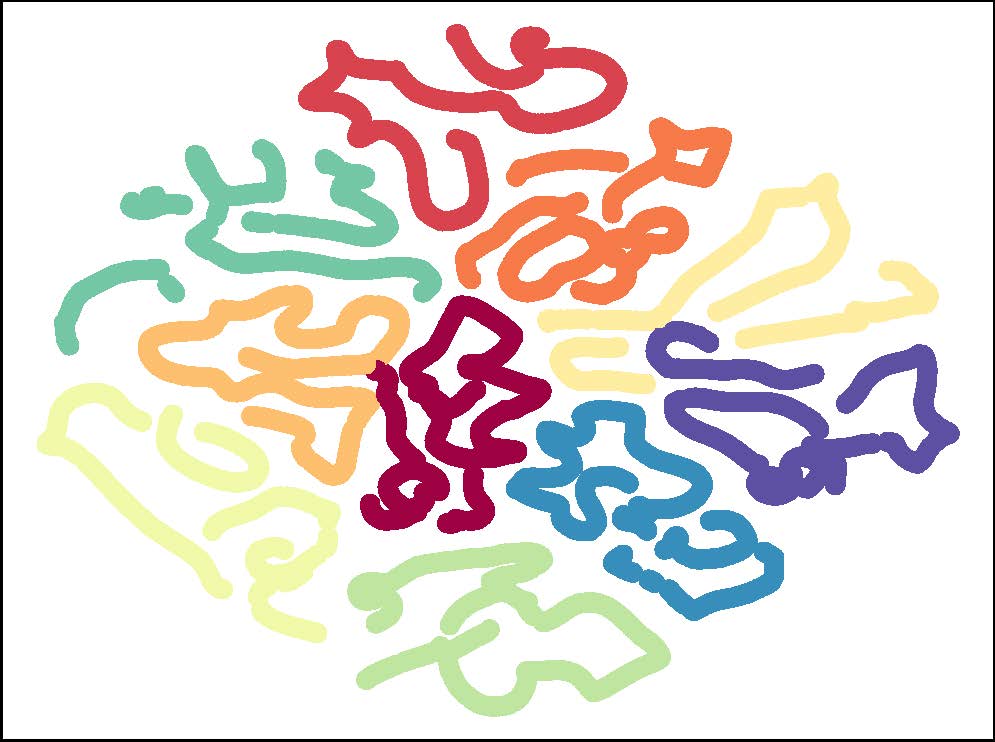}}}  
       \subfigure[Hi-MPC$^{h}$ (C)]{\scalebox{0.205}{\label{HiMPC_1_10_BIWI_P}\includegraphics[]{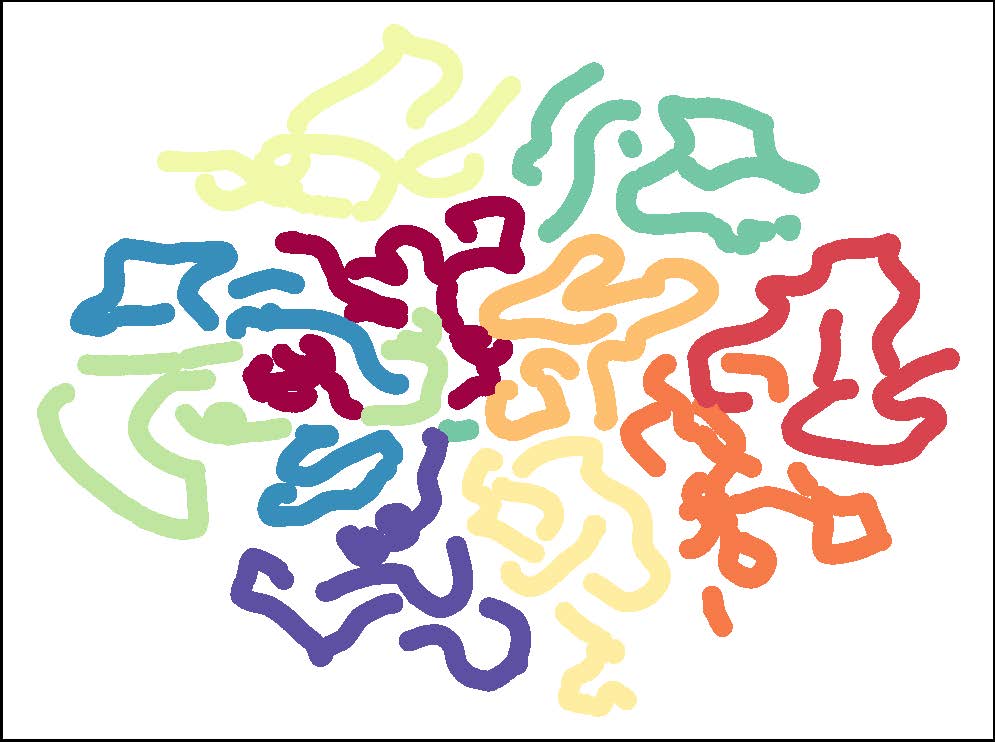}}} 
      \subfigure[Hi-MPC$^{h}$ (L)]{\scalebox{0.205}{\label{HiMPC_1_10_BIWI_B}\includegraphics[]{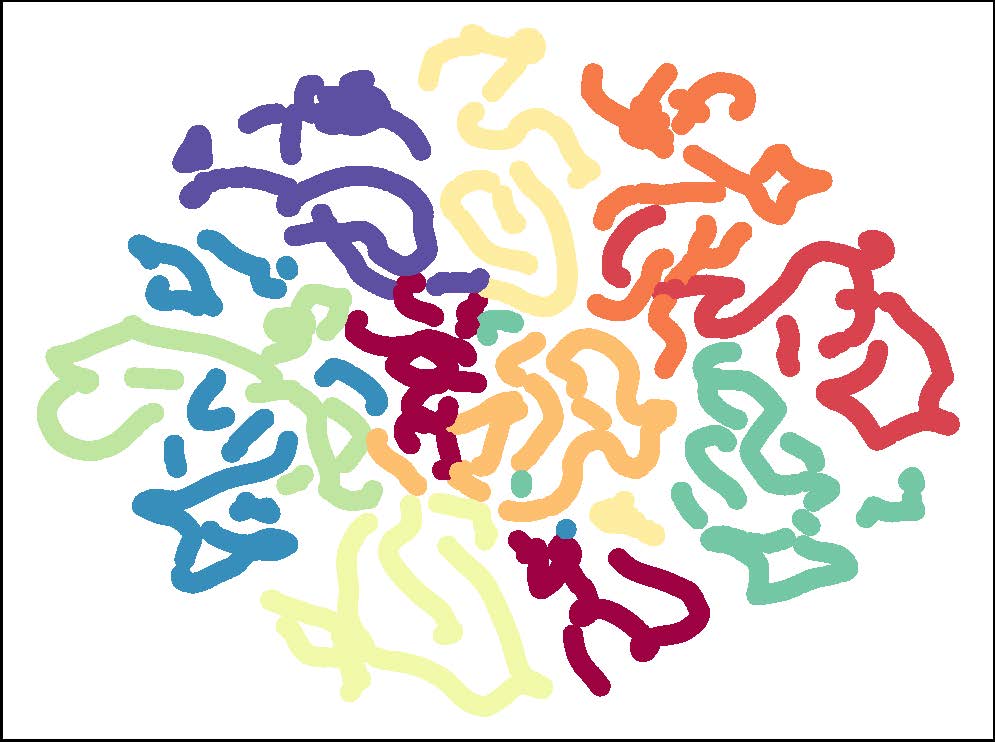}}}
    \caption{t-SNE visualization of skeleton representations learned from SM-SGE (a), SimMC (b), and Hi-MPC$^{h}$ ((c)-(f)) for the first 10 identities in BIWI. We visualize MSMR (M), joint-level (J), component-level (C), and limb-level representations (L) learned by Hi-MPC$^{h}$ in (c), (d), (e), and (f), respectively. Features of different identities are shown in different colors.}
    \label{TNS_comp}
\end{figure*}

\subsection{t-SNE Visualization of Skeleton Representations}
We provide a qualitative analysis with t-SNE \cite{van2008visualizing} visualization of skeleton representations, which are compared with two skeleton-based state-of-the-art methods, SM-SGE \cite{rao2021sm} and SimMC \cite{rao2022simmc}. As shown in Fig. \ref{HiMPC_1_10_BIWI}, the proposed MSMR learned from our approach achieves higher inter-identity separation than representations learned from SM-SGE. Compared with SimMC, our method is able to simultaneously learn coarse-to-fine ($e.g.,$ joint-level, component-level) skeleton representations with lower entropy, which forms evident identity groups in different levels, as shown in Fig. \ref{HiMPC_1_10_BIWI_J} and \ref{HiMPC_1_10_BIWI_P}. 
\hc{Such results not only suggest the effectiveness of  Hi-MPC$^{h}$ on learning discriminative representations ($i.e.,$ differences between ground-truth classes) from \textit{unlabeled} 3D skeleton data, but also demonstrate its ability on learning skeleton semantics ($e.g.,$ identity-specific patterns) at different levels, which is consistent with the conclusions in Sec. \ref{dis_levels}. However, it \rhc{can be} that the limb-level skeleton representations with the much more abstract body structure are more difficult to be clustered in the t-SNE visualization (see Fig. \ref{HiMPC_1_10_BIWI_B}). This implies that too much information loss of skeleton positions, structures, and dynamics ($e.g.,$ full dynamics of all joints) could reduce the model performance on learning recognizable pattern information of different identities.}


\section{Conclusion and Future Work}
\label{conclusion}
In this paper, we \rhc{proposed} a hierarchical skeleton meta-prototype contrastive learning (Hi-MPC) approach with a hard skeleton mining (HSM) mechanism to contrast and learn the most representative features from key informative skeletons for unsupervised person re-ID. The hierarchical representations of 3D skeletons are built to capture coarse-to-fine body features from different levels. To encourage more consistent contrastive learning to mine more representative skeleton prototypes, we propose to perform meta-transformation of instances and prototypes to contrast in multiple contrastive feature subspaces. We further devise a hard skeleton mining mechanism to assign larger importance to harder skeletons, so as to learn more valuable patterns and effective representations. Finally, we combine different level skeleton features learned from our approach to construct multi-level skeleton meta-representation (MSMR) for person re-ID. Our approach outperforms most state-of-the-art methods, and is also highly effective when applied to multi-view and RGB-based scenarios with estimated skeletons.

\mr{\textbf{Applicability.} Our model can be applied to either sensor-based skeleton data or RGB-estimated skeleton data. On the one hand, our approach can perform person re-ID using \textit{unlabeled} 3D skeleton data \textit{directly} captured from depth sensors such as Kinect. The quality of sensor-based skeleton data mainly depends on the precision of sensor algorithm ($e.g.,$ skeleton estimation algorithm of Kinect \cite{shotton2011real-time}) and capture distance ($i.e.,$ physical distance between sensor and subject). These factors influence the noise existed in the skeleton data and thus affect the performance of our model. 
On the other hand, when there are only cameras or images available, we can exploit state-of-the-art pose estimation models to first extract skeleton data from RGB images and then apply our model for person re-ID. 
The quality of RGB-estimated skeleton data highly relies on the precision of pose estimation models and the quality ($e.g.$, resolution) of the original RGB images, while their noise is typically larger than sensor-based skeletons using depth information. In this sense, the performance of our model can be further improved with higher-quality estimated skeletons.
}

\textbf{Limitation.} Despite achieving highly competitive performance on existing skeleton-based person re-ID benchmarks, our study has some limitations as follows. The 3D skeletons in this work are mainly collected from prevailing depth sensors such as Kinect \cite{shotton2011real-time} under controllable environments, while diverse skeleton collection ($e.g.,$ different devices in \mr{uncontrollable} environments) should be further studied for more general person re-ID scenarios. \mr{On the other hand, the person re-ID models under open-scenario ($e.g.$, long-distance, cross-device) skeleton data are not thoroughly investigated in this work, so more advanced sensor devices are expected to help collect such skeleton data with higher quality to validate model effectiveness.}
Compared with existing RGB-based person re-ID datasets ($e.g.,$ MSMT17), the scale of datasets \hc{used} in this study is relatively limited. Considering the \hc{lack} of large-scale 3D skeleton person re-ID benchmarks, we will collect and open more advanced and challenging skeleton datasets.


We believe that this work will promote the progress of general lightweight person re-ID models with 3D skeleton data, and there are several potential directions for future works. The hierarchical clustering can be further improved by aggregating key relational features among different level skeleton representations. Incorporating self-supervised pretext tasks ($e.g.,$ skeletal action prediction) into meta-prototype contrastive learning could encourage \hc{capture of} more valuable skeleton semantics for downstream tasks. More fine-grained hard skeleton mining mechanisms, such as ranking harder sequences or joints, can be devised to better guide the clustering and contrastive learning. Another potential direction is to explore diverse skeleton augmentation strategies to improve the sample capacity for higher-quality prototype learning. 
The generality of our model allows it to be transferred to various skeleton-based tasks, and it can hopefully synergize other data modalities \mr{($e.g.,$ depth/RGB images)} to drive more pattern recognition tasks. \mr{Considering the gaps existed in different data modalities and evaluation protocols ($e.g.,$ probe/gallery settings, single/multi-shot recognition), we will also explore a fair cross-modality evaluation and comparison protocol for comparing skeleton-based methods and different RGB/depth-based methods or multi-modal methods in our future work. }


\section*{Declarations}


\begin{itemize}
\item Funding: This research is supported by the National Research Foundation, Singapore under its AI Singapore Programme (AISG Award No: AISG2-PhD/2022-01-034[T]).
\item Conflict of interest:
The authors declare they have no
conflict of interest.
\item Availability of data, materials, and codes: All are available at \href{https://github.com/Kali-Hac/Hi-MPC}{https://github.com/Kali-Hac/Hi-MPC}.
\item Ethical statements: The datasets used in our work are officially shared by reliable research agencies, which guarantee that the collecting, processing, releasing, and using of data have gained the formal consent of \hc{participants}. To protect privacy, all individuals are anonymized with simple identity numbers. Our models and codes must only be used for legitimate research.
\end{itemize}

\bibliographystyle{IEEEtran}
\bibliography{sn-bibliography}




\end{document}